%% file: __main.tex
\documentclass{article}

\usepackage{include/jmlr2e}

\usepackage[utf8]{inputenc} % allow utf-8 input
\usepackage[T1]{fontenc}    % use 8-bit T1 fonts
\usepackage{hyperref}       % hyperlinks
\usepackage{url}            % simple URL typesetting
\usepackage{booktabs}       % professional-quality tables
\usepackage{amsfonts}       % blackboard math symbols
\usepackage{nicefrac}       % compact symbols for 1/2, etc.
\usepackage{microtype}      % microtypography
\usepackage{xcolor}         % colors

\usepackage{adjustbox}
\usepackage{graphicx}
\usepackage{subfig}

\usepackage{titletoc}
\usepackage{amsmath}
\usepackage{amssymb}
\usepackage{multirow}
\usepackage{array}
\usepackage{multicol}
\usepackage{tabularx}
\usepackage{comment}

\usepackage{algorithm}
\usepackage{algpseudocode}

\usepackage{hyperref}
\usepackage{cleveref}

\usepackage{wrapfig}
\usepackage{algpseudocode}
\usepackage{lipsum}

\usepackage{color}
\usepackage{colortbl}

\usepackage{enumitem}

\definecolor{orange1}{RGB}{246, 214, 184}
\definecolor{green1}{RGB}{183, 220, 177}

\input{macros}

\ShortHeadings{Embarassingly Simple Dataset Distillation}{}

\title{Embarassingly Simple Dataset Distillation}

\begin{document}

\author{
\centering \name Yunzhen Feng\footnotemark[1] \footnotemark[3] \quad \name Ramakrishna Vedantam\footnotemark[1] \quad \name Julia Kempe\footnotemark[1] \footnotemark[2] \\
        \addr \footnotemark[1] Center for Data Science, New York University   \\
        \addr \footnotemark[2] Courant Institue of Mathematical Sciences, New York University 
         \\
        \addr \footnotemark[3] yf2231@nyu.edu\\
        % \name Michael I.\ Jordan \email jordan@cs.berkeley.edu \\
        % \addr Division of Computer Science and Department of Statistics\\
        % University of California\\
        % Berkeley, CA 94720-1776, USA
        }

\maketitle

\setlength{\abovedisplayskip}{3pt}
\setlength{\belowdisplayskip}{3pt}

\input{0_abstract}

\input{1_intro}

\input{2_related_work}

\input{3_contribution}

\input{4_experiments}

\input{5_understanding}

\input{conclusion}

\clearpage

%\section*{References}

\bibliography{__main.bib}

\clearpage

\input{10_appendix}
%%%%%%%%%%%%%%%%%%%%%%%%%%%%%%%%%%%%%%%%%%%%%%%%%%%%%%%%%%%%

\end{document}

%% file: macros.tex
\newcommand{\rama}[1]{\textcolor{blue}{#1}}

\newcommand{\yunzhen}[1]{\textcolor{cyan}{Yunzhen: #1}}

\renewcommand{\paragraph}[1]{\noindent \textbf{#1}}

\newcommand{\pmvar}[1]{\scriptsize{$\pm$#1}}
\newenvironment{packed_itemize}{
\begin{list}{\labelitemi}{\leftmargin=2em}
\vspace{-6pt}
 \setlength{\itemsep}{0pt}
 \setlength{\parskip}{0pt}
 \setlength{\parsep}{0pt}
}{\end{list}}

%% file: 0_abstract.tex
\begin{abstract}
Dataset distillation extracts a small set of synthetic training samples from a large dataset with the goal of achieving competitive performance on test data when trained on this sample. In this work, we tackle dataset distillation at its core by treating it directly as a bilevel optimization problem. Re-examining the foundational back-propagation through time method, we study the pronounced variance in the gradients, computational burden, and long-term dependencies. We introduce an improved method: Random Truncated Backpropagation Through Time (RaT-BPTT) to address them. RaT-BPTT incorporates a truncation coupled with a random window, effectively stabilizing the gradients and speeding up the optimization while covering long dependencies. This allows us to establish new state-of-the-art for a variety of standard dataset benchmarks. A deeper dive into the nature of distilled data unveils pronounced intercorrelation. In particular, subsets of distilled datasets tend to exhibit much worse performance than directly distilled smaller datasets of the same size. Leveraging RaT-BPTT, we devise a boosting mechanism that generates distilled datasets that contain subsets with near optimal performance across different data budgets. 
\end{abstract}

%% file: 1_intro.tex
\section{Introduction}

Learning deep, overparameterized neural networks with stochastic gradient descent, backpropagation and large scale datasets has led to tremendous advances in deep learning. In practice, it is often observed that for a deep learning algorithm to be effective, a vast amount of training samples and numerous training iterations are needed.

In this work, we aim to explore the genuine necessity of vast training data and numerous training steps for achieving high test accuracy. To investigate, we limit the number of samples in the training set to be small (e.g., 1, 5, or 10 images per class) and the number of training steps to be small (e.g., on the order of 300 steps). This leads us to the concept of optimizing a small synthetic dataset, such that neural networks trained on this dataset perform well on the desired target distribution, a problem known as {\em Dataset Distillation} \citep{wang2018datasetdistillation}.

This is an instance of a bilevel optimization problem~\citep{DempeBO2020} where the output of one optimization problem (in this instance, the learning algorithm trained on the small dataset) is fed into another optimization problem (the generalization error on the target set) which we intend to minimize. In general, this problem is intractable, as the inner loop involves a multi-step computation with a large number of steps. Early works \citep{wang2018datasetdistillation,sucholutsky2021soft,deng2022remember} directly approached this problem via back-propagation through time (BPTT), unrolling the inner loop for a limited number of steps, before hitting an optimization bottleneck that called for alternative techniques. 
Later works have made steady progress by replacing the inner loop with closed-form differentiable {\em surrogates}, like the Neural Tangent Kernel \citep{nguyen2021kip,nguyen2021kipimprovedresults}, Neural Features \citep{zhou2022dataset} and Gaussian Process \citep{loo2022efficient}. This approach often requires looping thought a diverse pool of randomly initialized models during the optimization to alleviate the mismatch between the surrogate model and the actual one. Moreover, these approaches are limited to MSE loss; they tend to work better on {\em wider} models, where the surrogate approximation holds better, but give worse performance on the set of frequently used narrower models. Another line of works has modified the outer loop objective using {\em proxy training-metrics} like matching the trajectories of the network \citep{cazenavette2022dataset} or the gradients during training \citep{zhao2021datasetcondensation}. However, these methods either necessitate the storage of different trajectories or are impeded by subpar performance. %\yunzhen{These methods either require storing all checkpoints and unrolling or suffer from weak performance.} \julia{something is missing here!} either because the convex approximations of the inner loop do not account for feature learning of the data or because modified outer-loop metrics may not fully align with the test-loss objective. 
These observations lead to the question: Does there exist a simple and direct method for dataset distillation?

\begin{figure}[t]
\centering
\vspace{-10pt}
\begin{minipage}{.51\textwidth}
    \centering
    \includegraphics[width=\linewidth]{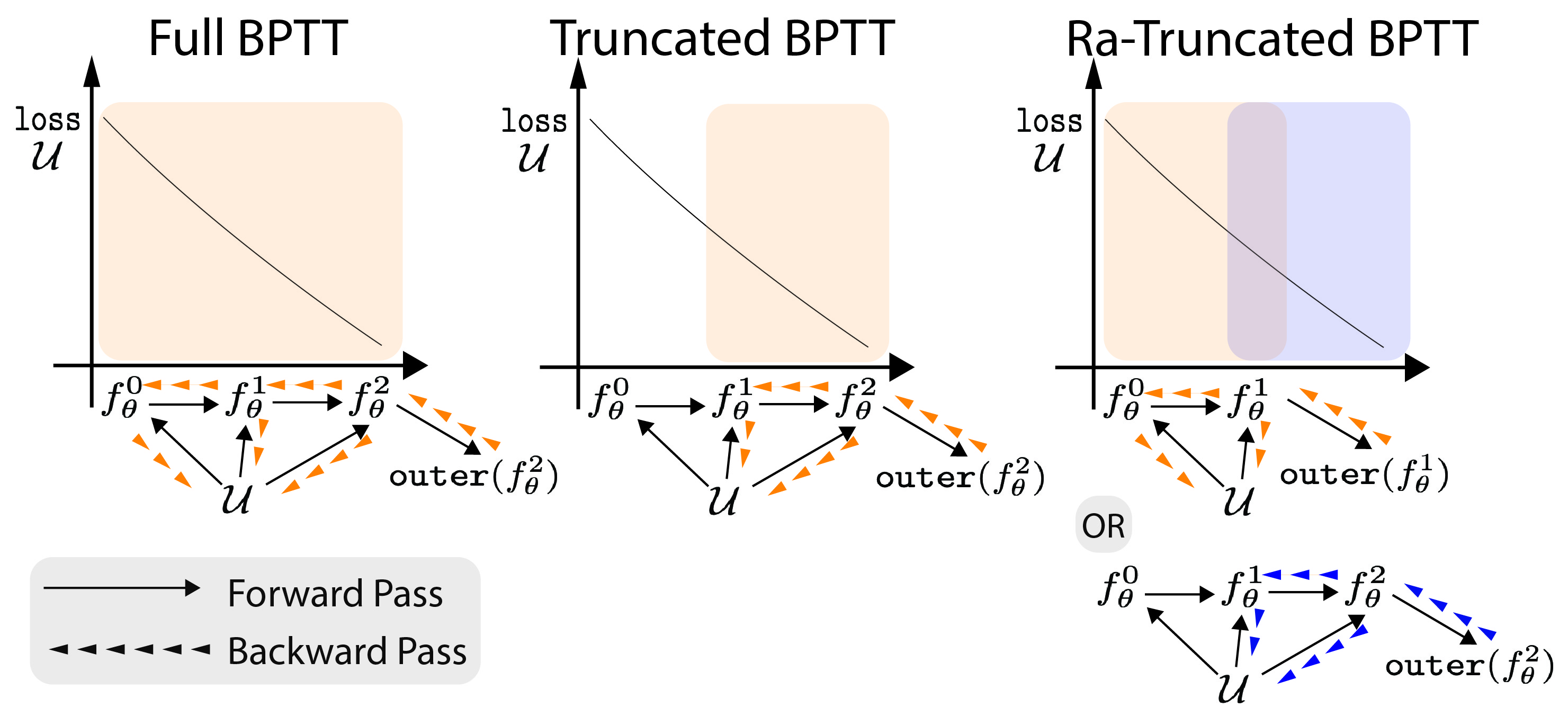}
    \caption{\textbf{Illustration of bilevel optimization} of the $\texttt{outer}$ loss when training for 3 steps. We show Full Backpropagation Through Time (BPTT) (left), Truncated BPTT (middle)
    and our proposed Randomized Truncated BPTT (right) (RaT-BPTT). RaT-BPTT picks a window in the learning trajectory (randomly) and tracks the gradients on the training dataset $\mathcal{U}$ for the chosen window, as opposed to T-BPTT that uses a fixed window, and BPTT that uses the entire trajectory.}
    \label{fig:teaser}
    \end{minipage} \hfill
    \begin{minipage}{.45\textwidth}
    \centering
    \vspace{-3pt}
    \includegraphics[width=\linewidth]{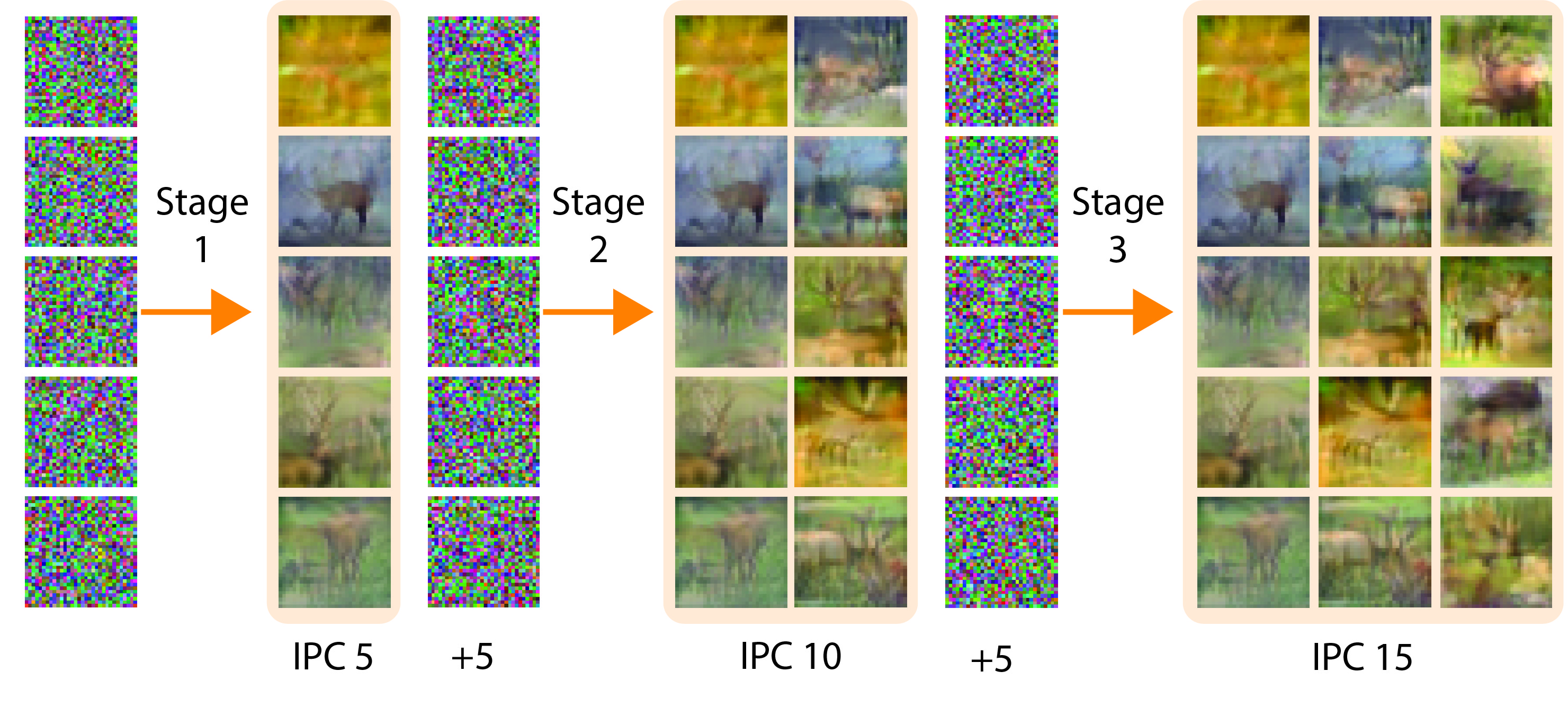}
    \caption{\textbf{Boosting Dataset Distillation (Boost-DD).} We start with
    5 randomly initialized images per class, distill the dataset into them (Stage 1) yielding five images per class (IPC5), then add five more random images and distill while reducing the learning rate on the first 5 (Stage 2) to yield IPC10, and so on, resulting in a nested dataset of different IPC. Boosting reduces higher order dependencies in the distilled datasets.}
    \label{fig:boosting-cartoon}
    \end{minipage}
    \vspace{-12pt}
\end{figure}

% In this paper, we adapt previous machinery for solving generic bilevel optimization problems via gradient based approaches to our problem, and demonstrate that one can achieve state-of-the-art performance across a vast majority of the CIFAR10, CIFAR100, CUB and TinyImageNet benchmarks. 

In this paper, we refine BPTT to address distinct challenges in dataset distillation, and achieve state-of-the-art performance across a vast majority of the CIFAR10, CIFAR100, CUB and TinyImageNet benchmarks. We start by re-examining BPTT, the go-to method for bi-level optimization problems \citep{finn17meta, lorraine20meta}. Notably, the inner problem of dataset distillation presents unique challenges -- the pronounced non-convex nature when training a neural network from scratch on the distilled data. One has to use long unrolling of BPTT to encapsulate the long dependencies inherent in the inner optimization. However, this results in BPTT suffering from slow optimization and huge memory demands, a consequence of backpropagating through all intermediate steps. This is further complicated by considerable instability in meta-gradients, emerging from the multiplication of Hessian matrices during long unrolling. Therefore, the performance is limited. 

To address these challenges, we integrate the concepts of randomization and truncation with BPTT, leading to the Random Truncated Backpropagation Through Time (RaT-BPTT) method. The refined approach unrolls within a randomly anchored smaller fixed-size window along the training trajectory and aggregates gradients within that window (see Figure~\ref{fig:teaser} for a cartoon illustration). The random window design ensures that the RaT-BPTT gradient serves as a random subsample of the full BPTT gradient, covering the entire trajectory, while the truncated window design enhances gradient stability and alleviates memory burden. Consequently, RaT-BPTT provides expedited training and superior performance compared to BPTT. 

Overall, our method is \emph{embarrassingly} simple -- we show that a careful analysis and modification of backpropagation lead to results exceeding the current state-of-the-art, without resorting to various approximations, a pool of models in the optimization, or additional heuristics. Since our approach does not depend on large-width approximations, it works for any architecture, in particular commonly used narrower models, for which methods that use inner-loop approximations perform less well. Moreover, our method can be seamlessly combined with prior methods on dataset re-parameterization \citep{deng2022remember}, leading to further improvements. To our knowledge, we are the first to introduce {\em truncated} backpropagation through time \citep{pmlr-v89-shaban19a} to the dataset distillation setting, and to combine it with {\em random} positioning of the unrolling window. %along the inner training trajectory. 

%\julia{We proceed with a modularization of RaT-BPP, also applicable to most other DD algorithms}
%\yunzhen{The transition here is too blunt. I would instead say we proceed to dissect current data...}
Having established the strength of our method, we proceed to dissect the structure of the learned datasets to catalyze further progress. In particular we address the observation, already made in prior work (e.g. \cite{nguyen2021kipimprovedresults}), that distilled data seems to show a large degree of {\em intercorrelation}. When training on a subset of distilled data, for instance 10 images per class extracted from a 50-image per class distilled dataset we observe a large degradation in test accuracy: the resulting dataset performs much worse than if it were distilled from scratch; even worse than training on a dataset of random train images of the same size! This property makes distilled data less versatile since for each desired dataset size we need to distill from scratch. To produce datasets that contain high performing subsamples, we propose {\em Boost-DD}, a boosting algorithm that produces a nested dataset without these higher-order correlations and only marginal performance loss. It works as a plug-and-play for essentially every existing gradient-based distillation algorithm (see Figure~\ref{fig:boosting-cartoon} for an illustration). 
%Through analysis of jointly distilled and boosted datasets we also shed light on whether these higher order correlations are necessary. 
To further our understanding of the learned dataset, 
we discuss the role of intercorrelation, as well as what information is captured in the distilled data through the lens of hardness metrics.
%\yunzhen{Moreover, we analyze what information is captured in the distillation using {\em hardness metrics}. This analysis sheds light on whether the higher order correlations are necessary and point to possible future improvements.}
%\julia{Move above before Overall, our contributions....and change this.}; in particular stemming from an analysis of the dataset using  and point to possible future improvements, with more details in the Appendix.

Overall, our contributions are as follows:
\vspace*{-5pt}
\begin{itemize}[leftmargin=10pt]
\setlength\itemsep{-1pt}
    \item \textbf{RaT-BPTT algorithm:} We propose RaT-BPTT by integrating truncation and randomization with BPTT, and achieves state of the art performance on various dataset distillation benchmarks. RaT-BPTT can be combined with data parametrization methods, leading to further improvements.
    %We show that a technique for bilevel optimization, truncated backpropagation, combined with randomization of the reverse-mode differentiation window, achieves state of the art performance on dataset distillation. \julia{You forgot a short piece on parametrization here.}
    \item \textbf{Boosting:} We propse a boosting-approach to dataset distillation (Boost-DD) that generates a modular synthetic dataset that contains nested high-performance subsets for various budgets. %We study higher-order interdependencies as we increase the number of samples.
    
    %We study the extent to which higher-order interdependencies might be bottlenecking the performance as we increase the number of samples.
    % \item \textbf{Hardness metrics:} We provide an analysis of the failure modes of current dataset
    % distillation approaches in terms of how they perform
    % on training data samples of various difficulty, to demonstrate that saturation occurs mostly on medium-hard samples. This opens the road to improving performance via hardness-pruning on the validation set.
\end{itemize}
This paper is structured as follows: Section \ref{sec:background} surveys prior work, Section \ref{sec:Methods} delineates and motivates our algorithm, RaT-BPTT, Section \ref{sec:experiments} presents experimental results and compares to prior art, and Section \ref{sec:understanding} details and evaluates our boosting algorithm. In Section 6 we summarize and discuss bottlenecks to further improvements.

%% file: 2_related_work.tex
\section{Background and Related Work}\label{sec:background}
 
 {\em Dataset Distillation}, introduced by \cite{wang2018datasetdistillation}, aims to condense a given dataset into a small synthetic version. When neural networks are trained on this distilled version, they achieve good performance on the original distribution. Dataset distillation shares many characteristics with {\em coreset selection} \cite{Jubran2019IntroductionTC}, which finds representative samples from the training set to still accurately represent the full dataset on downstream tasks.
 However, since dataset distillation generates synthetic samples, it is not limited to the set of images and labels given by the dataset and has the benefit of using continuous gradient-based optimization
techniques rather than combinatorial methods, providing added flexibility and performance.
Both coresets and distilled datasets have found numerous applications including speeding up model-training \cite{mirzasoleiman20a}, reducing catastrophic forgetting \cite{sangermano22ContLearn,zhou2022dataset},  federated learning \cite{hu2022fedsynth,song2022federated} and neural architecture search \cite{Such20aArchSearch}.

% For each unrolling, the gradient update shall be
% $$
% \theta_{i+1}(\mathcal{U}) = \theta_{i}(\mathcal{U}) - \alpha \frac{\partial \mathcal{L}(\theta_{i}(\mathcal{U}), \mathcal{U})}{\partial \theta}.
% $$

% \yunzhen{Julia: How should we posit the related work discussion here with the related work discussion in the introduction? I feel like the intro is still too long. I would suggest making either the intro shorter or this section shorter.}

Numerous follow up works have proposed clever strategies to improve upon the original direct bilevel optimization%,   like also learning soft labels \cite{bohdal2020flexible, sucholutsky2021soft} 
(see  \cite{lei2023survey, sachdeva2023survey, yu2023review, geng2023survey} for recent surveys and \cite{cui2022dc} for benchmarking). Yet, given the apparent intractability of the core bilevel optimization problem, most works have focused on 1) approximating the function in the inner-loop with more tractable expressions or 2) changing the outer-loop objective.
%and 3) re-parametrization of the data. Background on  3) can be found in Section \ref{sec:parametrization}. \julia{Honestly, let's just remove 3 here. It's not really a new method, since it can be combined with a variety of methods, no?} 

% Numerous follow up works have proposed clever strategies to improve upon the original direct bilevel optimization (Eq.~\eqref{eqn:bi_level_opt}) in \cite{wang2018datasetdistillation}, like also learning soft labels \cite{bohdal2020flexible, sucholutsky2021soft} (see  \cite{lei2023survey, sachdeva2023survey, yu2023review, geng2023survey} for recent surveys and \cite{cui2022dc} for benchmarking). 
% Most of them have focused on 1) approximating the function in the inner-loop with more tractable expressions, 2) changing the outer-loop objective and 3) re-parametrization of the data. 

{\em Inner-loop surrogates:} The first innovative works \cite{nguyen2021kip,nguyen2021kipimprovedresults} tackle inner-loop intractability by approximating the inner network with the Neural Tangent Kernel (NTK) which describes the neural net in the infinite-width limit with suitable initialization (\cite{JHG18,Lee+19,Aro+19b}) and allows for convex optimization, but scales unfavorably. To alleviate the scaling, random feature approximations have been proposed: 
 \cite{loo2022efficient}  leverage a Neural Network Gaussian process (NNGP) to replace the NTK, using MC sampling to approximate the averaged GP. \cite{zhou2022dataset} propose to use the Gram matrix of the feature extractor as the kernel, equivalent to only training the last layer with MSE loss. A very recent work \cite{loo2023dataset}  assumes that the inner optimization is convex
 by considering linearized training in the lazy regime
 and replaces the meta-gradient with implicit gradients, thus achieving most recent state-of-the-art. Yet all of these approaches inevitably face the discrepancies between learning in the lazy regime and feature learning in data-adaptive neural nets (e.g.~\cite{Ghorbani19featurelazy} and numerous follow ups) and often need to maintain a large model pool. Moreover, inner-loop surrogates, be it NTK, NNGP or random features, tend to show higher performance on {\em wide} networks, where the approximation holds better, and be less effective for the narrower models used in practice.
 
{\em Modified objective:} A great number of interesting works try to replace the elusive test accuracy objective with metrics that match the networks trained on full data and on synthetic data. \cite{zhao2021datasetcondensation} propose to match the gradient between the two networks with cosine similarity, with various variations (like differentiable data augmentation \citep{zhao2021differentiatble} (DSA)) and improvements \citep{jiang2022delving, lee2022dataset}. Other works pioneer feature alignment \citep{wang2022cafe}, matching the training trajectories (MTT, introduced in \cite{cazenavette2022dataset} and refined in \cite{du2023minimizing, cui2022scaling, zhang2022accelerating}), and loss-curvature matching \citep{shin2023lcmat}. However, it is unclear how 
well the modified outer-loop metrics align with the test-loss objective and most of these methods ends up with subpar performance.

%% file: 3_contribution.tex
\section{Methods}\label{sec:Methods}

In this section, we start by defining the dataset distillation problem to motivate our RaT-BPTT method. Denote the original training set as $\mathcal{D}$ and the distilled set as $\mathcal{U}$. With an initialization $\theta_0$ for the inner-loop learner $\mathcal{A}$, we perform the optimization for $T$ steps to obtain $\theta_{T}(\mathcal{U})$ with loss $\mathcal{L}(\theta_{T}(\mathcal{U}), \mathcal{D})$. We add $(\mathcal{U})$ to denote its dependence on $\mathcal{U}$. The dataset distillation problem can be formulated as

 \begin{equation}\label{eqn:bi_level_opt}
    \underset{\mathcal{U}}{\min} \,\, \mathcal{L}({\theta_T(\mathcal U)}, \mathcal{D}) \,\, \text{(outer loop)}\,\,\,\,\,\,\,
    \text{such that}\quad {\theta_T}(\mathcal U) = \mathcal{A}({\theta_0}, \mathcal{U}, T) \,\, \text{(inner loop)}
\end{equation}

%\subsection{Truncated Meta-Gradients}

The principal method for tackling bilevel optimization problems is  {\em backpropagation through time (BPTT) } in reverse mode. When the inner-loop learner $\mathcal{A}$ is gradient descent with learning rate $\alpha$, 
%within the inner loop with gradient descent and learning rate $\alpha$, 
we obtain the meta-gradient with respect to the distilled dataset by leveraging the chain rule:  
% $$\frac{\partial \mathcal{L}(\theta_{T}(\mathcal{U}), \mathcal{D})}{\partial \mathcal{U}} = \frac{\partial \mathcal{L}(\theta_{T}(\mathcal{U}), \mathcal{D})}{\partial \theta} \frac{\mathrm{d} \theta_{T}(\mathcal{U})}{\mathrm{d} \mathcal{U}}$$
%
% Let's decompose the second term: 
% \begin{equation}
%     \begin{aligned}
%         \frac{\mathrm{d} \theta_{T}(\mathcal{U})}{\mathrm{d} \mathcal{U}} &= \frac{\mathrm{d} \theta_{T-1}(\mathcal{U})}{\mathrm{d} \mathcal{U}} - \alpha \frac{\mathrm{d}}{\mathrm{d} \mathcal{U}}\frac{\partial \mathcal{L}(\theta_{T-1}(\mathcal{U}), \mathcal{U})}{\partial \theta} \\
%         &= \frac{\mathrm{d} \theta_{T-1}(\mathcal{U})}{\mathrm{d} \mathcal{U}} - \alpha \frac{\partial^2 \mathcal{L}(\theta_{i}(\mathcal{U}), \mathcal{U})}{\partial \theta^2} \frac{\mathrm{d} \theta_{T-1}(\mathcal{U})}{\mathrm{d} \mathcal{U}} - \alpha \frac{\partial^2 \mathcal{L}(\theta, \mathcal{U})}{\partial \theta_{T-1}(\mathcal{U}) \partial u} \\
%         &= \left[ 1 - \alpha \frac{\partial^2 \mathcal{L}(\theta_{T-1}(\mathcal{U}), \mathcal{U})}{\partial \theta^2} \right] \frac{\mathrm{d} \theta_{T-1}(\mathcal{U})}{\mathrm{d} \mathcal{U}} - \alpha \frac{\partial^2 \mathcal{L}(\theta_{T-1}(\mathcal{U}), \mathcal{U})}{\partial \theta \partial u} \\
%         &= - \alpha \sum_{i=1}^{T-1} \Pi_{j=i+1}^{T-1}\left[ 1 - \alpha \frac{\partial^2 \mathcal{L}(\theta_{j}(\mathcal{U}), \mathcal{U})}{\partial \theta^2} \right] \frac{\partial^2 \mathcal{L}(\theta_{i}(\mathcal{U}), \mathcal{U})}{\partial \theta \partial u}
%     \end{aligned}
% \end{equation}
%
% Therefore, the gradient is
\begin{equation}
    \mathcal{G}_{BPTT} = - \alpha \frac{\partial \mathcal{L}(\theta_{T}(\mathcal{U}), \mathcal{D})}{\partial \theta} \sum_{i=1}^{T-1} \Pi_{j=i+1}^{T-1}\left[ 1 - \alpha \frac{\partial^2 \mathcal{L}(\theta_{j}(\mathcal{U}), \mathcal{U})}{\partial \theta^2} \right] \frac{\partial^2 \mathcal{L}(\theta_{i}(\mathcal{U}), \mathcal{U})}{\partial \theta \partial u}
\end{equation}

The aforementioned computation reveals that the meta-gradient can be decomposed into $T-1$ parts. Each part essentially represents a matrix product of the form $\Pi[1-\alpha H]$ where every $H$ matrix corresponds to a Hessian matrix. Nonetheless, computing the meta-gradient demands the storage of all intermediate states to backpropagate through every unrolling step. This imposes a significant strain on GPU memory resources and diminishes computational efficiency. 

% \begin{figure}[tb]
%     \vspace*{-18pt}
%     \centering
%     \begin{subfigure}[t]{0.48\textwidth} % width of left subfigure
%         \centering
%         \includegraphics[width=\linewidth]{figs/final/gradients.pdf}   
%  %   \vspace*{-8pt}
%         %\refstepcounter{figure}
%         \captionsetup{labelformat=empty}
%         \caption[Figure \thefigure]{Meta-gradient norm in the first 500 steps. BPTT (unroll 120 steps) have unstable gradients. T-BPTT with unrolling 120 steps and backpropagating 40 steps stabilizes the gradient. For RaT-BPTT, for each epoch (25 batch-update steps) we randomly place the 40-step backpropagation window along the 120 unrolling. CIFAR10, IPC10.}
%         \label{fig:grad_comp}
%     \end{subfigure}
%     \hfill
%     \begin{subfigure}[t]{0.45\textwidth} % width of right subfigure
%         \centering
%         \includegraphics[width=\linewidth]{figs/final/bptt_acc.pdf}
%         % \refstepcounter{figure}
%         \caption[Caption for Figure \thefigure]{Test Accuracy during distillation with BPTT, T-BPTT, R-BPTT, and our RaT-BPTT. Using random unrolling (R-BPTT) and truncated window (T-BPTT) are both worse than BPTT-120. Combining them into RaT-BPTT gives the best performance. CIFAR10, IPC10. }
%         \label{fig:curr}
%     \end{subfigure}
% \end{figure}

\begin{figure}[tb]
\vspace*{-5mm}
\begin{minipage}{.48\linewidth}
    \centering
    \includegraphics[width=\linewidth]{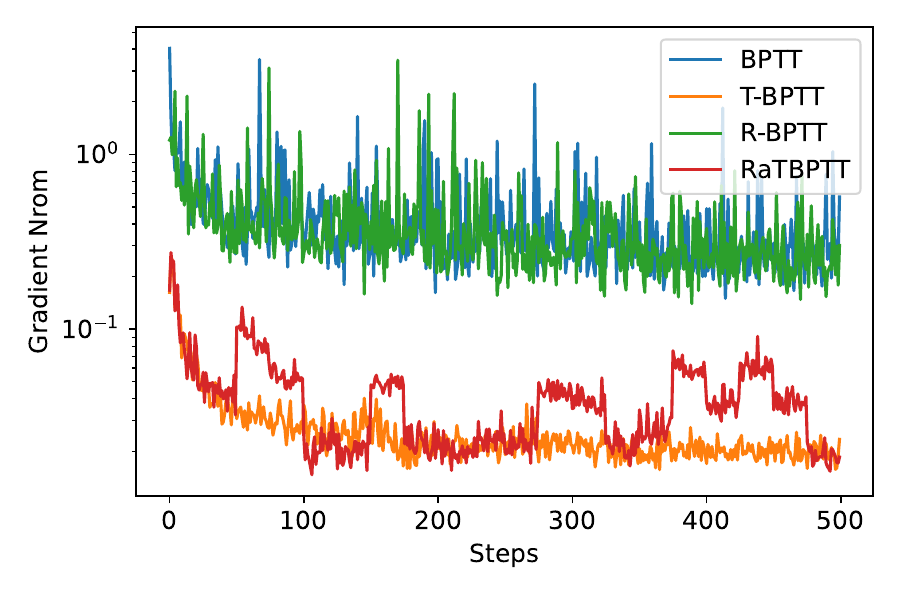}   
 %   \vspace*{-8pt}
    \caption{Meta-gradient norm in the first 500 steps. BPTT (unroll 120 steps) have unstable gradients. T-BPTT (unroll 120 steps and backpropagate 40 steps) stabilizes the gradient. For RaT-BPTT, for each epoch (25 batch-update steps) we randomly place the 40-step backpropagation window along the 120 unrolling. CIFAR10, IPC10.}
    \label{fig:grad_comp}
 %   \vspace{-5pt}
% \end{figure}
\end{minipage}\hfill
\begin{minipage}{.45\linewidth}
% \begin{figure}[htb]
%\vspace*{-21mm}
    \centering
    \includegraphics[width=\linewidth]{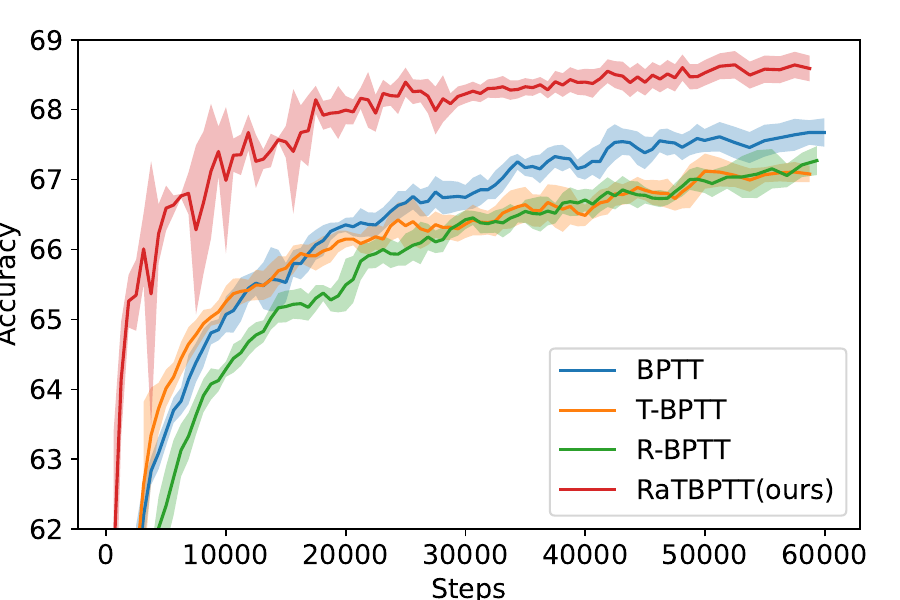}
    \caption{Test Accuracy during distillation with BPTT, T-BPTT, R-BPTT, and our RaT-BPTT. Using random unrolling (R-BPTT) and truncated window (T-BPTT) are both worse than BPTT. Combining them into RaT-BPTT gives the best performance. CIFAR10, IPC10. }
    \label{fig:curr}
\end{minipage}
    \vspace{-5pt}
\end{figure}
%    \vspace{-5pt}
% \end{figure}

% \begin{wrapfigure}{L}{\linewidth}
% % \vspace*{-23mm}
% \begin{minipage}{.45\linewidth}
% % \begin{figure}[htb]
% %\vspace*{-21mm}
%     \centering
%     \includegraphics[width=\linewidth]{figs/final/gradients.pdf}   
%  %   \vspace*{-8pt}
%     \caption{Meta-gradient norm in the first 500 steps. Both BPTT with 40 steps and 120 steps have unstable gradients. T-BPTT with unrolling 120 steps and backpropagating 40 steps stabilizes the gradient. For RaT-BPTT, for each epoch (25 batch-update steps) we randomly place the 40-step backpropagation window along the unrolling trajectory of length $\leq120$ and retain stability. CIFAR10, IPC10.}
%     \label{fig:grad_comp}
%  %   \vspace{-5pt}
% % \end{figure}
% \end{minipage}\hfill
% \begin{minipage}{.45\linewidth}
% % \begin{figure}[htb]
% %\vspace*{-21mm}
%     \centering
%     \includegraphics[width=\linewidth]{figs/final/bptt_acc.pdf}
%     \caption{Test Accuracy during distillation with BPTT, T-BPTT, R-BPTT, and our RaT-BPTT. Using random unrolling (R-BPTT) and truncated window (T-BPTT) are both worse than BPTT-120. Combining them into RaT-BPTT gives the best performance. CIFAR10, IPC10. }
%     \label{fig:curr}
% %    \vspace{-5pt}
% %\end{figure}
% %    \vspace{-5pt}
% % \end{figure}
% \end{minipage}
% \end{wrapfigure}

To circumvent these challenges, the prevalent strategy is {\em truncated} BPTT (T-BPTT) method \citep{williams1990efficient, puskorius1994truncated}, which unrolls the inner loop for the same $T$ steps but only propagates backwards through a smaller window of $M$ steps. In T-BPTT, the gradient is
\begin{equation}
    \mathcal{G}_{T-BPTT} = - \alpha \frac{\partial \mathcal{L}(\theta_{T}(\mathcal{U}), \mathcal{D})}{\partial \theta} \sum_{i=T-M}^{T-1} \Pi_{j=i+1}^{T-1}\left[ 1 - \alpha \frac{\partial^2 \mathcal{L}(\theta_{j}(\mathcal{U}), \mathcal{U})}{\partial \theta^2} \right] \frac{\partial^2 \mathcal{L}(\theta_{i}(\mathcal{U}), \mathcal{U})}{\partial \theta \partial u}
\end{equation}

The distinguishing feature of T-BPTT is its omission of the first $T-M+1$ terms in the summation; each omitted term is a product of  more than $M$ Hessian matrices. Under the assumption that the inner loss function is locally $\alpha-$strongly convex, \cite{pmlr-v89-shaban19a} shows that T-BPTT inherits convergence guarantees. The theoretical result comes from the diminishing contributions of the Hessian products. Strong convexity assumptions endow the Hessian matrices with positive eigenvalues. Consequently, $1-\alpha H$ will have all eigenvalues smaller than 1, and the product term $\Pi[1-\alpha H]$ vanishes as the number of factors increases. Therefore, T-BPTT could enjoy a similar performance compared with BPTT but with less memory requirement and faster optimization time. 

However, the inner task in our context diverges significantly from the realm of strong convexity. It contains training a neural network from scratch on the current distilled data with random initialization. This problem is intrinsically non-convex with multiple local minima. This beckons the question: how do BPTT and T-BPTT fare empirically?

\begin{wrapfigure}{rT}{0.57\textwidth}
\vspace*{-6mm}
\begin{minipage}{1\linewidth}
\begin{algorithm}[H]
    \caption{Dataset Distillation with RaT-BPTT. Differences from BPTT are {\color{purple} highlighted in purple.}}
    \label{alg:1}
    \textbf{Input:} Target dataset $\mathcal{D}$. T: total number of unrolling steps. M: truncated window size.
    \begin{algorithmic}[1]
    \State Initialize distilled data $\mathcal{U}$ from Gaussian
    \While{Not converged}
        \State {\color{purple} Uniformly sample N in $[M, T]$ as the current unrolling length}
        \State Sample a batch of data $d \sim \mathcal{D}$
        \State Randomly initialize $\theta_0$ from $p(\theta)$
            \For{$n=0\rightarrow N-1$}
            \State {\color{purple} If $n==N-M$, start accumulating gradients}
            \State Sample a mini-batch of distilled data $u_t\sim \mathcal{U}$
            \State Update network $\theta_{n+1} = \theta_n - \alpha \nabla \ell(u_n; \theta_n)$
            \EndFor
        \State Compute classification loss $\mathcal{L}= \ell(d, \theta_{N})$
        \State Update $\mathcal{U}$ with respect to $\mathcal{L}$.
    \EndWhile
    \end{algorithmic}
\end{algorithm}
\end{minipage}
\end{wrapfigure}

We visualize the training curve and the norm of meta-gradients through outer-loop optimization steps in Figure \ref{fig:curr}. The experiment is performed on CIFAR10 with IPC 10. A comparison between BPTT120 and T-BPTT reveals that: 1) The meta-gradients of BPTT manifest significantly greater instability than their T-BPTT counterparts. This observed volatility and norm discrepancy can be attributed to the omitted $T-M+1$ gradient terms. It underscores the highly non-convex nature of the inner problem, characterized by Hessian matrices with negative eigenvalues. The compounded effects of these negative eigenvalues amplifies the variance from different initializations, creating the unstable gradient behavior. With the gradient stabilized, T-BPTT achieves faster improvement during the initial phase. 
2) BPTT ends up with higher accuracy than T-BPTT. This indicates that important information from the initial phase is disregarded in T-BPTT — a notable concern given that the rapid optimization of neural networks usually happens during the early stage of the inner loop. The challenge thus is how to harmonize the good generalization performance of BPTT with the computational speedup of T-BPTT.

To this end, we propose the {\em Random} Truncated BPTT (RaT-BPTT) in Algorithm \ref{alg:1}, which randomly places the truncated window along the inner unrolling chain. The gradient of RaT-BPTT is 

\begin{equation}
    \mathcal{G}_{RaT-BPT T} = - \alpha \frac{\partial \mathcal{L}(\theta_{N}(\mathcal{U}), \mathcal{D})}{\partial \theta} \sum_{i=N-M}^{N-1} \Pi_{j=i+1}^{N-1}\left[ 1 - \alpha \frac{\partial^2 \mathcal{L}(\theta_{j}(\mathcal{U}), \mathcal{U})}{\partial \theta^2} \right] \frac{\partial^2 \mathcal{L}(\theta_{i}(\mathcal{U}), \mathcal{U})}{\partial \theta \partial u}
\end{equation}

Looking at the gradients, RaT-BPTT differs by randomly sampling M consecutive parts in $\mathcal{G}_{BPTT}$ and leaving out the shared Hessian matrix products. Therefore, RaT-BPTT is a subsample version of BPTT, spanning the entire learning trajectory. Moreover, the maximum number of Hessians in the product is restricted to less than than M. It thus inherits the benefits of both the accelerated performance and gradient stabilization from T-BPTT. As illustrated in Figure \ref{fig:curr}, RaT-BPTT consistently outperforms other methods throughout the optimization process. We also examine performing full unrolling along trajectories of randomly sampled lengths (R-BPTT) as a sanity check. The gradients are similarly unstable and the performance is worse than full unrolling with BPTT. We further provide an ablation study in Section \ref{sec:ablation} on the necessity of having a moving truncated window and the rationale of random uniform sampling in Algorithm \ref{alg:1}.

%\begin{wrapfigure}{r}{0.4\textwidth}
%\vspace{-25pt}
%\centering
%    \includegraphics[width=0.9\linewidth]{figs/final/bptt_acc.pdf}
%    \caption{Test Accuracy during distillation with BPTT, T-BPTT, R-BPTT, and our RaT-BPTT. Using random unrolling (R-BPTT) and truncated window (T-BPTT) are both worse than BPTT-120. Combining them into RaT-BPTT gives the best performance. CIFAR10, IPC10. }
%    \label{fig:curr}
%    \vspace{-5pt}
%\end{wrapfigure}

%\subsection{Other Modification}

%% file: 4_experiments.tex
\section{Experimental Results}\label{sec:experiments}

In this section, we present an evaluation of our method, RaT-BPTT, comparing it to a range of SOTA methods across multiple benchmark datasets. 

\textbf{Datasets} We run experiments on four standard datasets, CIFAR-10 (10 classes, $32\times 32$), CIFAR-100 (100 classes, $32\times 32$, \cite{krizhevsky2009learning}), Caltech Birds 2011 (200 classes, CUB200, $32\times 32$, \cite{wah2011caltech}) and Tiny-ImageNet (200 classes, $64\times 64$, \cite{le2015tiny} ). We distill datasets with 1, 10, and 50 images per class for the first two datasets and with 1 and 10 images per class for the last two datasets.

\textbf{Baselines} We compare our methods to the first two lines of works as we discussed in related work (Section \ref{sec:background}), including 1) \textit{inner-loop surrogates}: standard BPTT (the non-factorized version of LinBa in \citep{deng2022remember}), Neural Tangent Kernel (KIP) \citep{nguyen2021kipimprovedresults}, Random Gaussian Process (RFAD) \citep{loo2022efficient}, and empirical feature kernel (FRePO) \citep{zhou2022dataset}, and reparameterized convex implicit gradient (RCIG) \citep{loo2023dataset}, 2) \textit{Modified objectives}: gradient matching with augmentation (DSA) \citep{zhao2021differentiatble}, distribution matching (DM) \citep{zhao2023distribution}, trajectory matching (MTT) \citep{cazenavette2022dataset}, and flat trajectory distillation (FTD) \citep{cui2022scaling}. The works on parametrization \citep{deng2022remember, liu2022dataset, kim2022dataset} are  complementary to our optimization framework and can be combined with RaT-BPP for improved performance, as we illustrate for the SOTA case of linear basis \citep{deng2022remember} in Section \ref{sec:parametrization}. 
%Consequently, we have chosen not to include a direct comparison within the main body of this work. A comprehensive comparison, however, can be found in the Appendix.

\textbf{Setup} Building upon existing literature, we employ standard ConvNet architectures \citep{zhao2021differentiatble, deng2022remember, cazenavette2022dataset} —three layers for $32\times 32$ images and four layers for $64\times 64$ images. Our distilled data is trained utilizing Algorithm 1, with the Higher package \citep{grefenstette2019generalized} aiding in the efficient calculation of meta-gradients. We opt for a simple setup: using Adam for inner optimization with a learning rate of 0.001, and applying standard augmentations (flip and rotation) on the target set. Parameters such as unrolling length and window size are determined via a validation set.

\textbf{Evaluation}
During the evaluation phase, we adhere to the standard augmentation protocol as per \cite{deng2022remember, zhao2021differentiatble}. We conduct evaluations of each distilled data set using ten randomly selected neural networks, reporting both the mean and standard deviation of the results. For all other baseline methodologies, we record the best value reported in the original paper. Note that \cite{zhou2022dataset, loo2023dataset} employs a 4 or 8 times wider ConvNet to reduce discrepancies between surrogate approximations and actual training. To ensure alignment with this protocol, we provide a transfer evaluation of our method, that is we distill with a narrow network and evaluate with a wide network. We also re-evaluate their checkpoints (when available) for narrow networks using their code. Complete details, along with links to our code and distilled checkpoints, can be found in the Appendix.

\setlength{\tabcolsep}{4.5pt}
\vspace{12pt}
\begin{table*}[hb]
\caption{\small Performance of different dataset
distillation techniques on standard datasets. %\textbf{IPC\{X\}} denotes the performance for X images per class (IPC). 
The
\textbf{AVG} column denotes the average performance across
all the other columns. * denotes works where performance evaluated with wider ConvNets. FRePO and RCIG are the re-evaluated results with narrow networks. \colorbox{orange1}{Results} denotes the best results for narrow networks while \colorbox{green1}{results} denotes best for wide networks.}
\begin{adjustbox}{width=\columnwidth,center} 
\centering
\begin{tabular}{cl|ccc|ccc|cc|cc|c}
\toprule
\multicolumn{2}{c|}{Dataset} & \multicolumn{3}{c|}{CIFAR-10} & \multicolumn{3}{c|}{CIFAR-100} & \multicolumn{2}{c|}{CUB200} & \multicolumn{2}{c|}{T-ImageNet} & \multirow{2}{*}{AVG}\\
\multicolumn{2}{c|}{Img/class(IPC)}      & 1 & 10 & 50 & 1 & 10 & 50 & 1 & 10 & 1 & 10 \\
\midrule
%Baseline & Random & & 34.3\pmvar{0.9} & 51.2\pmvar{0.6} & 4.2\pmvar{0.3} & 14.6\pmvar{0.5}\\
%\midrule
 & BPTT \citep{deng2022remember} & 49.1\pmvar{0.6 } & 62.4\pmvar{0.4} & 70.5\pmvar{0.4} & 21.3\pmvar{0.6} & 34.7\pmvar{0.5} & - & - & - & - & - & -\\
& KIP* \citep{nguyen2021kipimprovedresults} &  49.9\pmvar{0.2}  &  62.7 \pmvar{0.3} & 68.6\pmvar{0.2} & 15.7\pmvar{0.2} & 28.3\pmvar{0.1} &  - & - & - & - & - & -\\
\multirow{1}{*}{Inner}& RFAD* \citep{loo2022efficient}& 53.6\pmvar{1.2} & 66.3\pmvar{0.5} & 71.1\pmvar{0.4} & 26.3\pmvar{1.1} & 33.0\pmvar{0.3} &  - & - & - & - & - & - \\
\multirow{1}{*}{Loop} & FRePO* \citep{zhou2022dataset} & 46.8\pmvar{0.7} & 65.5\pmvar{0.6} & 71.7\pmvar{0.2} & 28.7\pmvar{0.1} & 42.5 \pmvar{0.2} &  44.3\pmvar{0.2}  & 12.4\pmvar{0.2}   &  16.8\pmvar{0.1}  & 15.4\pmvar{0.3}  & 25.4\pmvar{0.2} & 36.9\pmvar{0.3}\\
& FRePO & 45.6\pmvar{0.1} & 63.5\pmvar{0.1} & 70.7\pmvar{0.1} & 26.3\pmvar{0.1} & 41.3 \pmvar{0.1} &  41.5\pmvar{0.1}  & -  &  -  & 16.9\pmvar{0.1}  & 22.4\pmvar{0.1} & - \\
& RCIG* \citep{loo2023dataset} & \cellcolor{green1}53.9\pmvar{1.0} & 69.1\pmvar{0.4} & 73.5\pmvar{0.3} & \cellcolor{green1}39.3\pmvar{0.4} & 44.1\pmvar{0.4} & 46.7\pmvar{0.1} & 12.1\pmvar{0.2} & 15.7\pmvar{0.3} & \cellcolor{green1}25.6\pmvar{0.3} & \cellcolor{green1}29.4\pmvar{0.2} & 40.9\pmvar{0.4}\\
& RCIG  & 49.6\pmvar{1.2} & 66.8\pmvar{0.3} & - & \cellcolor{orange1}35.5\pmvar{0.7} & - & - & - & - & \cellcolor{orange1} 22.4\pmvar{0.3} &  -  & - \\
\midrule
 & DSA \citep{zhao2021differentiatble} & 28.8\pmvar{0.7} & 52.1\pmvar{0.5} & 60.6\pmvar{0.5} & 13.9\pmvar{0.3} & 32.3\pmvar{0.3} & 42.8\pmvar{0.4} & 1.3\pmvar{0.1} & 4.5\pmvar{0.3} & 6.6\pmvar{0.2} & 14.4\pmvar{2.0} & 25.7\pmvar{0.7} \\
\multirow{1}{*}{Modified} & DM \citep{zhao2023distribution} & 26.0\pmvar{0.8} & 48.9\pmvar{0.6} & 63.0\pmvar{0.4} & 11.4\pmvar{0.3} & 29.7\pmvar{0.3} & 43.6\pmvar{0.4} & 1.6\pmvar{0.1} & 4.4\pmvar{0.2} & 3.9\pmvar{0.2} & 12.9\pmvar{0.4} & 24.5\pmvar{0.4}\\
\multirow{1}{*}{Objectives}& MTT \citep{cazenavette2022dataset} & 46.3\pmvar{0.8} & 65.3\pmvar{0.7} & 71.6\pmvar{0.2} & 24.3\pmvar{0.3} & 40.1\pmvar{0.4} & 47.7\pmvar{0.3} & 2.2\pmvar{0.1} & - & 8.8\pmvar{0.3} & 23.2\pmvar{0.2} & -\\
& FTD \citep{du2023minimizing} & 46.8\pmvar{0.3} & 66.6\pmvar{0.3} & 73.8\pmvar{0.2} & 25.2\pmvar{0.2} & 43.4\pmvar{0.3} & \cellcolor{orange1}50.7\pmvar{0.3} & - & - & 10.4\pmvar{0.3} & \cellcolor{orange1}24.5\pmvar{0.2} & -\\
\midrule
& Ours &  \cellcolor{orange1} 53.2\pmvar{0.7} & \cellcolor{orange1}69.4\pmvar{0.4}  &  \cellcolor{orange1}75.3\pmvar{0.3} &  \cellcolor{orange1}35.3\pmvar{0.4} & \cellcolor{orange1}47.5\pmvar{0.2}  & \cellcolor{orange1} 50.6\pmvar{0.2} &  \cellcolor{orange1}13.8\pmvar{0.3} & \cellcolor{orange1}17.7\pmvar{0.2} & 20.1\pmvar{0.3} & \cellcolor{orange1}24.4\pmvar{0.2} & \cellcolor{orange1}40.8\pmvar{0.3} \\
& {Ours (transfer to wide)} &  \cellcolor{green1} 54.1\pmvar{0.4} & \cellcolor{green1}71.0\pmvar{0.2}  & \cellcolor{green1}75.4\pmvar{0.2} & 36.5\pmvar{0.3}   &  \cellcolor{green1}47.9\pmvar{0.2} &  \cellcolor{green1}51.0\pmvar{0.3} & \cellcolor{green1}14.2\pmvar{0.3} & \cellcolor{green1} 17.9\pmvar{0.3} & 20.3\pmvar{0.1} & 24.9\pmvar{0.1} & \cellcolor{green1}41.2\pmvar{0.3}\\
\bottomrule
\end{tabular}
\end{adjustbox}
\label{tab:sota_results_table}
\end{table*}

\subsection{Benchmark Performance}

Our simple approach to dataset
distillation demonstrates competitive
performance
across a number of datasets (Table \ref{tab:sota_results_table}). With
10 and 50 images per class, our approach
gets the state-of-the-art (SOTA)
results on the CIFAR-100, CIFAR-10, and
CUB200 datasets (Table \ref{tab:sota_results_table} last two rows). Moreover, we achieve
these results without any  approximations
to the inner loop. When considering all IPC values in $\{1, 10, 50\}$, across all
datasets, our approach performs as well as the RCIG method up to statistical significance. Encouragingly,
even though our bilevel optimization is not biased towards wider networks, we obtain state-of-the-art performance
even for wide networks on CIFAR10, CIFAR100, and CUB200
across all IPC values. Moreover,
when the datasets from the wider-network
approaches are evaluated on practical,
narrower settings we find that they
show a significant drop in performance, 
going from 39.3\% to 35.5\% (RCIG, CIFAR100, IPC1) or from 25.4\% to 22.4\% (FrePO, TinyImageNet, IPC10). Thus, our work generalizes
gracefully to wider networks that are used by previous
work (improving in performance) as well as narrower
networks (for which we can tune  directly). This
is a significant advantage of our work over prior state-of-the-art.

%Overall, we notice that RaT-BPTT
%achieves state-of-the-art or close
%to state-of-the-art in \cite{loo2023dataset} based on their released checkpoints. 
\begin{comment}
We are SOTA on CIFAR10 all settings, CIFAR100 (50.5 vs 50.7), TinyImageNet IPC1, CUB200 both. Our method could be further combined with parameterization methods, for example \cite{deng2022remember}. We already surpass all the results from parameterization papers \cite{deng2022remember, liu2022dataset, kim2022dataset} except IPC1 CIFAR10. We include the comparison with reparameterization in the appendix.

\yunzhen{[Things need to be addressed]}
\begin{itemize}
    \item \yunzhen{Whether we keep the MIT paper's results here? \cite{loo2023dataset}. They use wide network and achieve significant results for TinyImageNet IPC50, 29\% with wide network} 
    \item \yunzhen{There is another paper that I ignored in CVPR 2023, \cite{zhang2022accelerating}. They build upon a parameterization method. We could ignore.}
    \item \yunzhen{Shall we discard the result for TinyImagenet or shall we ignore the MIT paper?}
    \item We would delete the comparison with parameterization methods.
    \item 
\end{itemize}
\end{comment}

% KIP & RFAD & FRePo & DSA & MTT & TESLA & Ours 

\subsection{Combination with Parametrization Methods}\label{sec:parametrization}

A separate and complimentary line of work aims to improve the optimization via parameterization of the distilled dataset. \cite{liu2022dataset, wang2023dim} leverage encoder-decoders,  \cite{lee2022kfs, cazenavette2023glad} use generative priors, \cite{kim2022dataset, liu2023dream} propose multi-scale augmentation, and \cite{deng2022remember} designs linear basis with weights for the dataset.

\begin{wrapfigure}{r}{0.6\linewidth}
    \centering
    % \vspace{-12pt}
    \begin{minipage}{\linewidth}
% \begin{table*}
\caption{Combination of RaT-BPTT with linear parameterization leads to further improvement. We only present those settings where parameterization outperforms the standard RaT-BPTT.}
\begin{adjustbox}{width=\columnwidth,center} 
\centering
\begin{tabular}{cl|ccc|ccc}
\toprule
\multicolumn{2}{c|}{Dataset} & \multicolumn{2}{c}{CIFAR-10}  \\
\multicolumn{2}{c|}{Img/class(IPC)}      & 1 & 10  \\
\midrule
Para- & IDC \citep{kim2022dataset} & 50.0\pmvar{0.4} & 67.5\pmvar{0.5} \\
meteri-& LinBa \citep{deng2022remember} & 66.4\pmvar{0.4} &  71.2\pmvar{0.4}  \\
zation & HaBa \citep{liu2022dataset} & 48.3\pmvar{0.8} &  69.9\pmvar{0.4}  \\
\midrule
& Linear + RaT-BPTT &  \cellcolor{orange1} 68.2\pmvar{0.4} & \cellcolor{orange1} 72.8\pmvar{0.4}  \\
\bottomrule
\end{tabular}
\end{adjustbox}
\label{tab:params}

% \end{table*}
\end{minipage}
% \vspace{-5pt}
\end{wrapfigure}

Note that our performance improvements come from a careful study of the bilevel optimization problem. In principle, RaT-BPTT is complimentary to most of these parameterization ideas and can be seamlessly intergrated. For instance, we adopt the linear basis from \cite{deng2022remember} within our framework.%, regardless of label learning usage \julia{???}. 
We only study the case of CIFAR10, as for all the other benchmarks RaT-BPP gives better performance even without data parametrization. Without any hyper-parameter tuning, we can improve their performance by around 1.6\%, leading to astonishing numbers of 68.2\% for IPC1 and 72.8\% for IPC10 (the numbers w/o parameterization are 53.2\% and 69.4\% respectively). The results are shown in Table \ref{tab:params}. 
We leave the exploration of other combinations to future work. %\julia{why did you cut the nice table?}

% \begin{adjustbox}{width=\columnwidth,center} 
% \centering
% \begin{tabular}{cc|ccc|ccc}
% \toprule
% \multicolumn{2}{c|}{Dataset} & \multicolumn{3}{c|}{CIFAR-10} & \multicolumn{3}{c}{CIFAR-100} \\
% \multicolumn{2}{c|}{Img/class(IPC)}      & 1 & 10 & 50 & 1 & 10 & 50 \\
% \midrule
% \multirow{2}{*}{Parameterization} & IDC \citep{kim2022dataset} & 50.0\pmvar{0.4} & 67.5\pmvar{0.5} & 74.5\pmvar{0.1} & - & 44.8\pmvar{0.2} & - \\
% & LinBa \citep{deng2022remember} & 66.4\pmvar{0.4} &  71.2\pmvar{0.4} & 73.6\pmvar{0.5} & 34.0\pmvar{0.4} & 42.9\pmvar{0.7} & -  \\
% & HaBa \citep{liu2022dataset} & 48.3\pmvar{0.8} &  69.9\pmvar{0.4} & 74.0\pmvar{0.2} & 33.4\pmvar{0.4} & 40.2\pmvar{0.2} & 47.0\pmvar{0.2} \\
% \midrule
% & Linear + RaT-BPTT &  \cellcolor{orange1} 68.2\pmvar{0.4} & \cellcolor{orange1} 72.8\pmvar{0.4}  &   &   &  & \\
% \bottomrule
% \end{tabular}
% \end{adjustbox}
% \label{tab:params}

% \end{table*}

\subsection{Ablations on the Random Truncated Window}\label{sec:ablation}

\begin{wrapfigure}{r}{0.45\textwidth}
\vspace{-17pt}
%   \centering
   \includegraphics[width=0.9\linewidth]{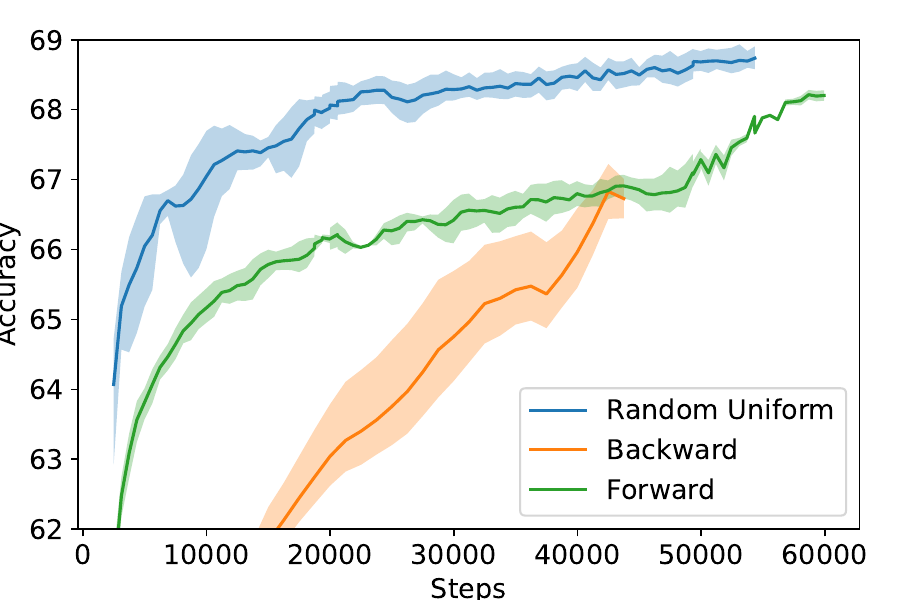}
   \vspace{-8pt}
    \caption{Comparison between random uniform truncation, backward moving, and forward moving. Random uniform truncation gives the best performance across the whole training process. N=120, T=40 for IPC10 with CIFAR10.}
    \label{fig:curr_comp}
     \vspace{-15pt}
\end{wrapfigure}

In Section \ref{sec:Methods}, we justify the necessity of performing truncation to speed up and stabilize the gradient, and the necessity of changing the truncated window to cover the entire trajectory. Now we provide an ablation study on how to select the truncated window. We compare three methods, 1) random uniform truncation, 2) backward moving, and 3) forward moving. For the forward (backward) moving method, we initialize the window at the end (beginning). It is then shifted forward (backward) by the window size whenever the loss remains stagnant for 2,000 steps.

From Figure \ref{fig:curr_comp}, it is surprising that randomly uniform window placement achieves the best performance across the whole training process. A closer examination of the forward and backward moving curves suggests that altering the window's positioning can spur noticeable enhancements in accuracy. Such findings reinforce the idea that distinct truncation windows capture varied facets of knowledge, bolstering our intuition about the need for a comprehensive trajectory coverage by the window.

One might ask whether uniform sampling is the best design. Actually the answer is no. With careful tuning by sampling more on the initial phase, we find that one can further improve the final accuracy by 0.4\% for CIFAR10 with IPC10. However, it introduces an additional hyper-parameter that requires careful tuning. To keep our method simple, we choose to go with the uniform one.

%% file: 5_understanding.tex
\section{Intercorrelations and Boosting}\label{sec:understanding}

\begin{figure}[tb]
\centering
\vspace{-10pt}
    \begin{minipage}[t]{0.48\linewidth}
        \centering
        \includegraphics[width=1.0\linewidth]{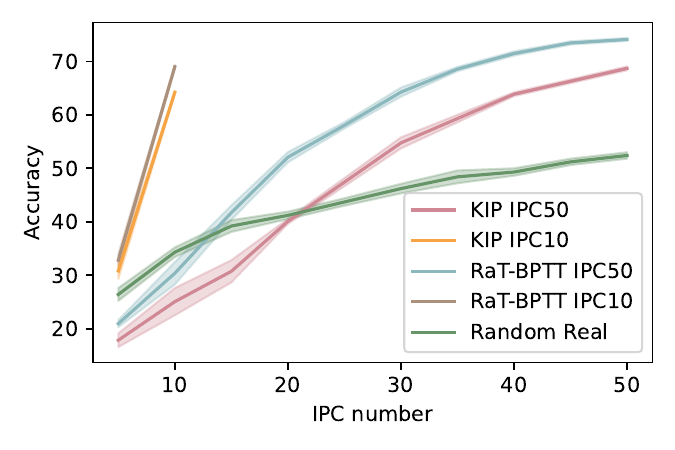}
        \caption{Test performance of random subsamples of IPC50 distilled dataset from KIP and our method for various sample sizes, compared to performance of real randomly sampled data of the same size. CIFAR10.}
        \label{fig:comp_subsample}
    \end{minipage}
    \hfill
    \begin{minipage}[t]{0.44\linewidth}
        \centering
        % \vspace{-0.3mm}
        \includegraphics[width=\linewidth]{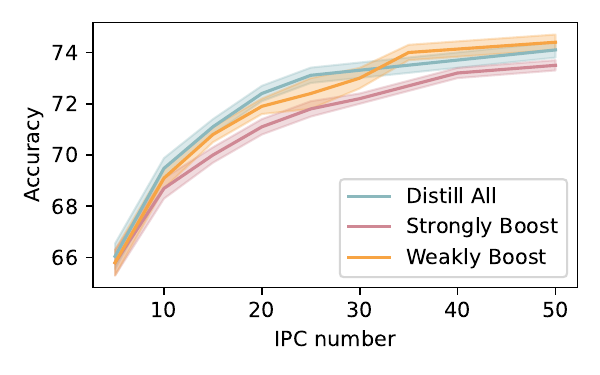}
        \caption{\small Performance of fully distilled, strongly and weakly boosted distilled data. Boosting essentially retains performance compared to jointly distilled data. %First block subsamples of strongly boosted data perform optimally. 
    CIFAR10, RaT-BPTT.}
        \label{fig:boostDD}
    \end{minipage}
    \vspace{-5pt}
\end{figure}

Having established the strength of our method, we proceed to dissect the structure of the learned datasets to catalyze further progress. Nearly all the current distillation method optimize the data jointly. Such joint optimization often leads to unexpected behavior which is largely absent in the original dataset. For instance, \cite{nguyen2021kipimprovedresults} observe that subsampling a large distilled dataset leads to significantly low performance. The data is jointly learned, and therefore can be {\em correlated}. A particularly noticeable consequence of the correlation is the necessity to re-distill the dataset for different distillation budgets, with minimal opportunity to leverage previous efforts.
In Figure \ref{fig:comp_subsample} , we compare subsampling accuracies of subsamples of various sizes coming from an IPC50 dataset produced with KIP (and RaT-BPTT). We validate previous observation and show its pervalance for various IPC settings. Specifically, we see that subsampling 5 images per class from an IPC50 distilled dataset not only gives performance way below an IPC5 dataset learned from scratch, but performs even worse than training on 5 {\em random} training images per class. Compared with KIP, our approach, RaT-BPP, demonstrates a reduced degradation in performance, particularly at intermediate IPC values, possibly attributable to its independence from approximations of the entire inner optimization loop. Still, while subsamples from RaT-BPP-distilled data fairs better than those from KIP-distilled data, for smaller sample sizes (up to IPC10) they fare worse than random samples. Is there potential for further improvement? Can we distill datasets containing subsets that lead to good accuracy for downstream tasks, comparable to the one achieved when we distill smaller datasets directly?

\begin{wrapfigure}{r}{0.55\textwidth}
\vspace*{-12pt}
\begin{minipage}{0.55\textwidth}
\begin{algorithm}[H]
    \caption{Boosted Dataset Distillation (Boost-DD)}
    \label{alg:bdd}
    \textbf{Input:} Target dataset $\mathcal{D}$. Distillation-Algorithm $\mathcal{A}$ with initiatlization procedure $\mathcal{I}(size)$ and meta-learning rate $\alpha$, outputting distilled data $\mathcal{U}=\mathcal{A}(\mathcal{D},\mathcal{I}(size))$ ($|\mathcal{U}|=size$). Block size $b$. Number of blocks $J$. Boosting-strength $\beta \in [0,1]$.\\
    \textbf{Output:} Distilled data $\mathcal{U}$ with $|\mathcal{U}|=b\cdot J$.
    \begin{algorithmic}[1]
    \State{Distill first block of size $b$: $\mathcal{U}_0:=\mathcal{A}(\mathcal{D},\mathcal{I}(b))$.}
    \For{$j=1 \ldots J-1$ }
        \State{Distill increased data $\mathcal{U}_j=\mathcal{A}(\mathcal{D},\mathcal{U}_{j-1} \cup \mathcal{I}(b))$} using ``stale" meta-learning rate $\alpha_s=\beta \cdot \alpha$ on the first $(j-1)\cdot b$ data points and $\alpha$ on the last $b$.
        \EndFor
   \State{$\mathcal{U}:=\mathcal{U}_{J-1}$.}
    \end{algorithmic}
\end{algorithm}
\end{minipage}
% \vspace*{-4mm}
\end{wrapfigure}

To address this challenge, we propose {\em boosted} dataset distillation (Boost-DD) (in Algorithm \ref{alg:bdd}). Boost-DD controls inter-correlation within distilled data and can be integrated with any gradient-based dataset distillation algorithm. The central idea is to construct the distilled data iteratively with smaller data groups called "blocks" as illustrated in Figure~\ref{fig:boosting-cartoon}. For IPC50, we divide all images into blocks of IPC5. We start from a distilled IPC5. Each time, we  add another fresh block of IPC5, and optimize the new block with reduced learning rate on the existing blocks by a factor of $\beta \leq 1$. The extreme case where the learning rate is zero for previous blocks is termed {\em strongly}-boost ($\beta=0$). The advantage of this approach is that initial blocks remain unchanged when the new block is added, ensuring consistent performance. This results in a "nested" dataset where subsampling earlier blocks yields high-performing subsets! We call the algorithm {\em weakly}-boost  $\beta$ is non-zero and perform experiments with $\beta=0.1$.

Figure \ref{fig:boostDD} shows how weakly-boost and strongly-boost compare to distilling each of the sub-datasets from scratch. It is important to highlight that the curve for `strongly-boost' represents the accuracy curve obtainable when subsampling across different budgets. We observe that even  strongly-boost results in exceedingly minor sacrifice in performance compared to joint distillation from scratch, especially with larger distilled datasets. In the Appendix we present a visual comparison of images distilled jointly with images distilled with boosting. Boosted images seem to show larger diversity and seem closer to real images.

%% file: conclusion.tex
\section{Discussion}

% \subsection{Inner-correlation}

% \subsection{Which data have been learnt}

In this work, we proposed a simple yet effective method for dataset distillation, based on random truncated backpropagation through time. Through a careful analysis of BPTT, we show that randomizing the window allows to cover long dependencies in the inner problem while  truncation addressed the unstable gradient and the computational burden. Our  method achieves state of the art performance across multiple standard benchmarks, across both narrow as well as wide networks. Nonetheless, several design choices are guided by intuitions and observations, leaving room for improvement. We defer a detailed limitation discussion to Appendix \ref{sec:limitation}.

\begin{figure}[tb]
\centering
\vspace{-8mm}
    \begin{minipage}{0.48\linewidth}
        \centering
        % \vspace{6mm}
        \includegraphics[width=1\linewidth]{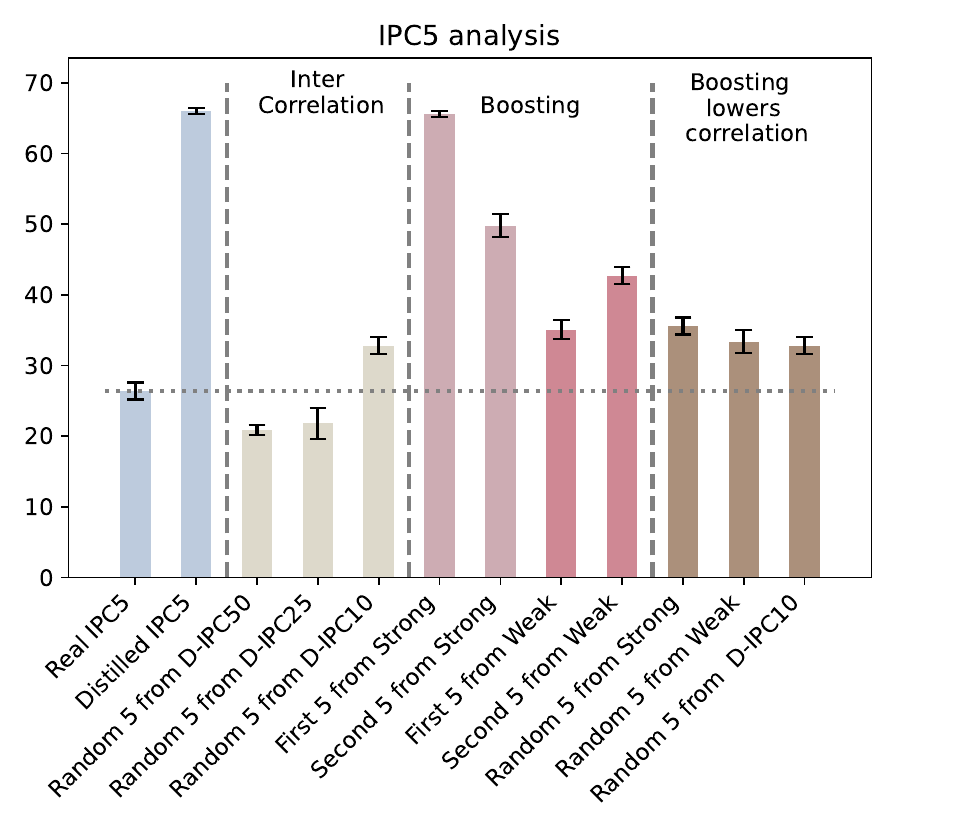}   
    \caption{Performance of subsamples compared to fully distilled and random real images for IPC5: y-axis shows test accuracy a) Random IPC5 from IPC50 and IPC25 performs worse than random \textit{real} IPC5, indicating the strong inter-correlation learned in the dataset. b) IPC5 building blocks of boosting perform quite well. c) Random IPC5 from boosted IPC10 performs better than random IPC5 from standard IPC10. Boosting somewhat lowers the inter-correlation.}
    % \vspace{-5mm}
    \label{fig:ipc5}
    \end{minipage}\hfill
    \begin{minipage}{0.48\linewidth}
        \centering
        \vspace{6pt}
        \includegraphics[width=1\linewidth]{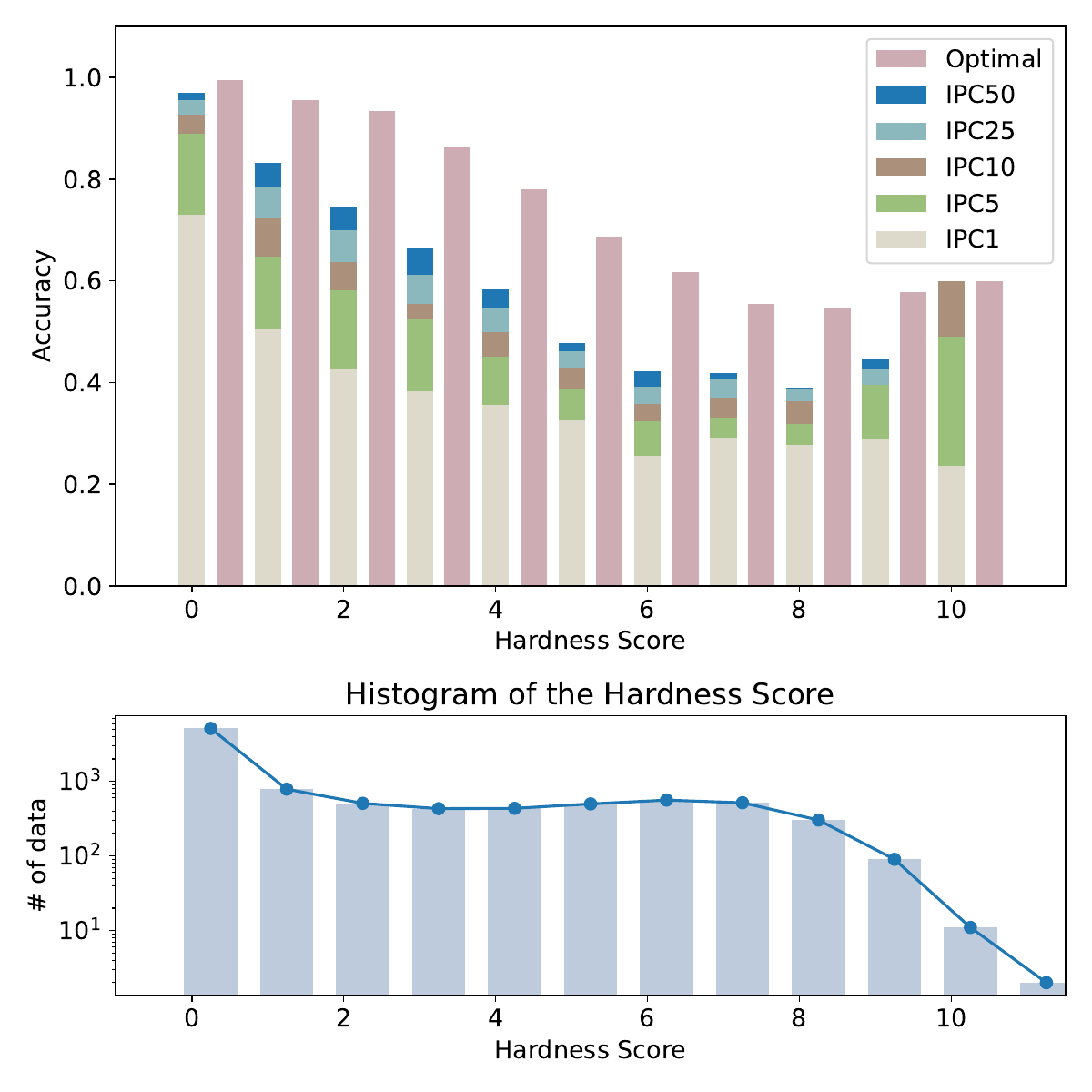}
    % \vspace{-6pt}
    \caption{\textbf{Top:} Hardness score vs accuracy of dataset distillation (for various images per class (IPC)). ``Optimal'' indicates accuracy when training on the entire 
    real data. \textbf{Bottom:} Histogram of the hardness scores. Notice how score 0 (unforgettable) examples are distilled well (top), but harder
    examples %(score > 0) 
    are progressively harder to distill.
    Details in Appendix \ref{sec:hardness}}
    \label{fig:hardness}
    \end{minipage}
    \vspace{-8pt}
\end{figure}

Further, we address the catastrophic degradation in performance of subsets of distilled data
%alleviate the inner correlation problem of the distilled dataset 
with our boosting method. It allows us to  create a single versatile dataset for various distillation budgets with minimal performance degradation. However, the boosted dataset still has inner correlation between blocks. %As observed in Figure \ref{fig:ipc5}, blocks distilled earlier influence the optimization of subsequently distilled blocks. \yunzhen{Previous sentence is obvious even without the experimental results.} 
This is evident in Figure \ref{fig:ipc5} when comparing the performance of the first IPC5 block with the second one obtained via strongly-boost (though both of them are much higher than sampling a random IPC5 from the jointly distilled IPC50). 
Moreover, as shown in Figure \ref{fig:boostDD}, weakly-boost for larger IPC eventually outperforms joint training. Since weakly-boost generates less inter-correlated datasets, this hints at the possibility that strong intercorrelations are one reason for diminishing returns observed when increasing the size of distilled datasets. 
While higher-order correlations may potentially encode more information, they also compromise data interpretability, diverging from the standard IID paradigm. Is it possible to further minimize these correlations, especially in larger IPC datasets?  We leave these questions to future research.

We also attempt to understand other factors bottlenecking the gains when scaling up distilled data. Specifically, we try to understand what information is learnt in the distilled data by dissecting the accuracy on evaluation samples. We leverage a hardness score that characterizes whether data is easy or hard, stratify the accuracy 
by hardness and compare it for the original network and the network trained on distilled data for a range of IPC in Figure~\ref{fig:hardness} (details in Appendix \ref{sec:hardness}). 
One would have hoped that adding more distilled data would help to distill more of the hardness tail, but this is not the case. This suggests that future work might benefit from focusing on how one can distill data that is better adapted to larger hardness scores, for instance by infusing validation batches with harder data, placing emphasis on the middle of the distribution. A preliminary promising study is presented in Appendix \ref{sec:hardness}.

\section{Acknowledgements}

This work was supported by the National Science Foundation under NSF Award 1922658. This work was also supported in part through the NYU IT High Performance Computing resources, services, and staff expertise. YF would like to thank Di He, Weicheng Zhu, Kangning Liu, and Boyang Yu for discussions and suggestions.

%% file: 10_appendix.tex
\appendix

% \julia{Do we want to switch C and D?}

\section{Limitations}\label{sec:limitation}

\paragraph{Algorithm Design} The design of our method is primarily guided by intuitions and observations from empirical studies. Throughout the algorithm's development, we aim to strike a balance between scalability and effectiveness. Our approach currently involves tuning the ratio between the unrolling length and window size, scaling the unrolling length in accordance with the IPC number and GPU size. While this approach has demonstrated promise, we acknowledge that the current algorithmic choice might not represent the absolute optimal solution. Further research could investigate alternative algorithm designs.

\paragraph{Application to larger models and datasets} A notable strength of our methodology is its versatility: it is compatible with all differentiable loss functions and network architectures, emphasizing its broad applicability. However, we only focus on illustrating the method's capabilities with standard benchmarks in the literature. This decision leaves a promising avenue for future work to apply and validate our method across various domains and tasks beyond image classification. It's also worth highlighting that while surrogate-based techniques are constrained to using the MSE loss to convexify the inner problem, our approach is agnostic to the specific loss function employed. This flexibility paves the way for our method's application in other realms, such as audio and text data distillation.
 
\paragraph{GPU memory usage} Despite the significant improvements introduced by RaT-BPTT, it still necessitates unrolling and backpropagating over several steps, which require storing all intermediate parameters in the GPU. Consequently, this method incurs substantial memory consumption, often exceeding that of directly training the model. For larger models, one might need to implement checkpointing techniques to manage memory usage effectively. 

% \textbf{Sampler and boosting limitations} The boosting process can reduce inner correlation by training each block consecutively, but it also tends to increase the overall training time relatively. Exploring novel strategies to effectively lower inner correlation could be a promising research direction. Besides, we only employ the forgetting score to understand which data points are difficult to distill. One could use other selection method for analysis and the adaptive training. 

\section{Other Related Work}

In this section, we discuss further works related to dataset distillation or hardness metrics.

%{\em coreset selection} Dataset distillation provides an alternative to coreset selection \cite{Jubran2019IntroductionTC}, which finds representative samples from the training set to still accurately represent the full dataset on downstream tasks. Dataset distillation shares many characteristics with coresets, however, instead of selecting subsets of the training data distillation generates synthetic samples. It is thus not limited to the set of images and labels given by the dataset and has the benefit of using continuous gradient-based optimization techniques rather than combinatorial methods, providing added flexibility and performance. 

{\em Boosting}: It is noteworthy that \cite{liu2023slimmable} has also identified challenges associated with retraining distilled datasets for varying budgets. Their proposed solution adopts a top-down approach, aiming to slim a large distilled dataset. In contrast, our method follows a bottom-up strategy, producing a modular dataset designed to accommodate various budgets. Moreover, one of our motivations is to address and study the intercorrelation problem.

{\em Hardness metrics:}
One way to study generalization performance of neural nets is to understand which data points are ``easy'' or ``hard'' to learn for a neural network. There is an intimate relationship to {\em data pruning} which tries to understand and quantify which subsets of the data can be pruned with impunity, while maintaining the performance of the neural net when trained on the remainder\footnote{The boundary between {\em coresets} and {\em data pruning} is fluid; the former term is used for small subsets of the training set, while the latter usually refers to removing only a fraction of the training data, like $25\%$ \cite{paul2021deep}.}. 
Inspired by the phenomenon of catastrophic forgetting, \cite{toneva2018forgetting} are the first to study how the learning process of different examples in the dataset varies. In particular, the authors analyze whether some examples are harder to learn than others (examples that are forgotten and relearned multiple times through learning.) and define a {\em forgetting score} to quantify this process. To our knowledge, our work is the first to use this tool to understand how learning on distilled data differs from learning on full data, to identify bottlenecks. The idea to ``enrich'' the validation data during the data distillation process appears in \cite{liu2023dream}, who chose more ``representative'' data to learn from, as determined by  k-means clustering. To our knowledge, we are the first to propose learning from ``harder-to-learn'' data towards more efficient data distillation.

{\em Extensions of Dataset Distillation:} Beyond the conventional dataset distillation formula that aims to minimize generalization error, there have been advances in optimizing metrics for various objectives. These include dataset generation tailored for generalization attacks \cite{YuWu21}, adversarial perturbations \cite{TsilivisKempe22}, and generating distilled data with an emphasis on robustness \cite{tsilivis2022can}.

\section{Experiments}
\subsection{Experimental Details}

% \julia{Please fix indentation! Either make it paragraph or add noindent before each one.}

\noindent \textbf{Data Preprocessing}: Leveraging a regularized ZCA transformation with a regularization strength of $\lambda=0.1$ across all datasets, our approach adheres to the methods established by prior studies \cite{nguyen2021kip, nguyen2021kipimprovedresults, zhou2022dataset, loo2022efficient, deng2022remember}. We apply the inverse ZCA transformation matrix for distillation visualization, using the mean and standard deviation to reverse-normalize optimized data.

\noindent \textbf{Models} Following previous works, we use Instance Normalization for all networks for both training and evaluation.

\noindent \textbf{Initialization} In contrast to conventional real initialization widely used in nearly all previous works, we employ random initialization for distilled data, hypothesizing that there is a reduction in bias from such uninformative initialization. Data are initialized via a Gaussian distribution and normalized to norm 1. For RaT-BPTT, we note comparable performance and convergence between random and real initialization.

\noindent \textbf{Label Learning} Following previous works that leverage learnable labels, we optimize both the data and label for CIFAR10-IPC50, all IPCs for CIFAR100, CUB-200, and Tiny-ImageNet. We forego normalization for label probability, hence the labels retain their positive real value representation.

\noindent \textbf{Training} In addition to the RaT-BPTT algorithm, we incorporate meta-gradient clipping with an exponential moving average to counter gradient explosion. We find that the proper combination of normalizing initialization and learning rate (0.001 for Adam) is crucial for successful distillation image training. While using instance normalization, an image scaled by $\alpha$ leads to meta-gradient scaling by $\frac{1}{\alpha}$. As a result, one should use an $\alpha$ times larger learning rate for Adam or $\alpha^2$ times larger for SGD to achieve the same optimization trajectory. We thus adopt a similar initialization scale to that of neural network training (normalized to norm 1), combined with a standard learning rate of 0.001 when using Adam. To maintain meta-gradient stability, we employ batch sizes of 5,000 for CIFAR-10 and CIFAR-100, 3,000 for CUB-200, and 1,000 for Tiny-ImageNet. Note that one should aim to further increase the batch size for Tiny-ImageNet until all the GPU memory is used.

\noindent \textbf{Hyperparameters} In an effort to minimize tuning requirements, we adhere to a standard baseline across all configurations. Specifically, we utilize the Adam optimizer for both the inner loop (network unrolling) and the outer loop (distilled dataset optimization) with learning rates uniformly set to 0.001. We refrain from applying weight decay or learning rate schedules that are used in prior works \cite{zhou2022dataset, loo2023dataset}. 

\noindent \textbf{Evaluation} We evaluate our optimized data using a seperate held-out test dataset (the test set in the corresponding dataset). We adopt the same data augmentation as in previous work \cite{deng2022remember}. For depth 3 convolutional networks, we train using Adam with a learning rate of 0.001. No learning rate schedule is used.

\noindent \textbf{Code and Checkpoints} The code and checkpoints for RaT-BPTT could be found at \url{https://github.com/fengyzpku/Simple_Dataset_Distillation}

\subsection{Ablations on curriculum}

\begin{figure}[htb]
    \centering
    \includegraphics[width=0.45\linewidth]{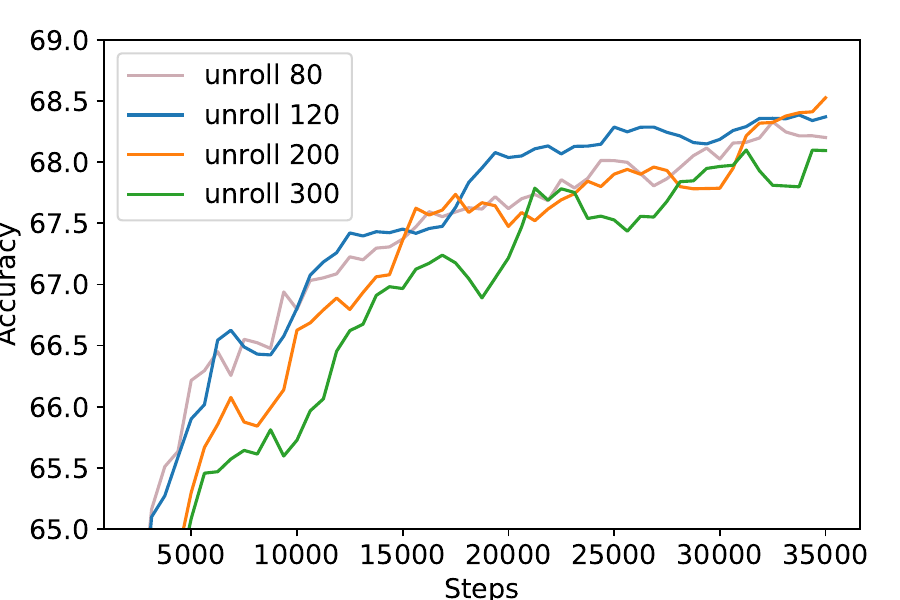}
    \includegraphics[width=0.45\linewidth]{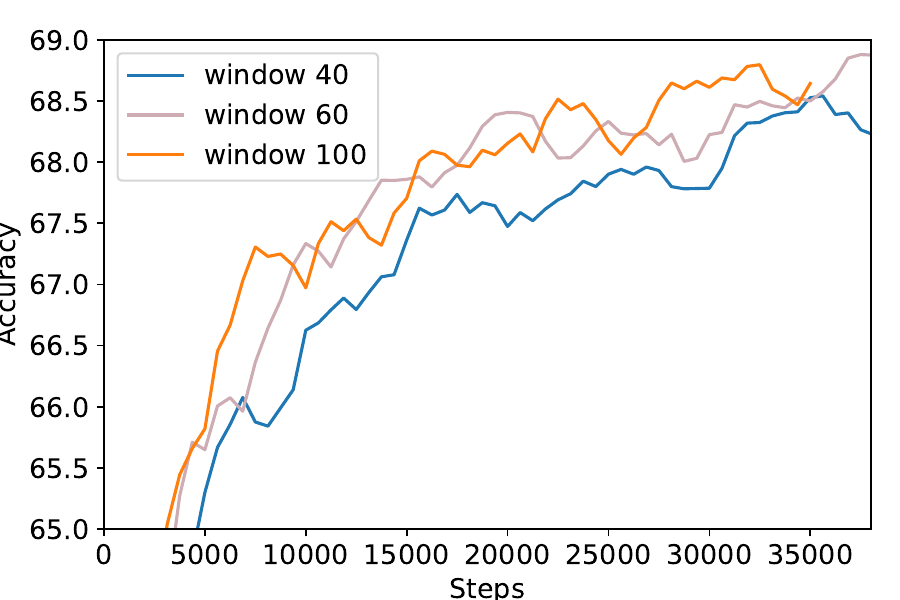}
    \caption{\textbf{Left} Test accuracy during distillation for different unrolling length of 80, 120, 200, 300 with fixed window size 40. CIFAR-10, IPC10. \textbf{Right} Test accuracy during distillation for different window size in 40, 60, 100 with fixed unrolling length 200. }
    \label{fig:diff_unroll}
\end{figure}

Our RaT-BPTT implementation hinges on tuning two hyperparameters: unrolling length and backpropagation window size. This section presents an ablation study exploring these parameters for CIFAR-10, IPC10.

\noindent \textbf{Unrolling length}

We initially fix the window size at 40 while varying the unrolling length. Notably, unrolling length governs the long-term dependencies we can capture within the inner loop. Figure \ref{fig:diff_unroll} reveals that a moderate unrolling length, between twice and five times the window size, yields similar performance. However, disproportionate unrolling, as seen with a window size of 40 and unrolling length of 300, detrimentally affects performance.

\noindent \textbf{Window size}

Next, we fix the unrolling length at 200 and experiment with window sizes of 40, 60, and 100. Figure~\ref{fig:diff_unroll} shows the latter two sizes yield comparable performance. In RaT-BPTT, GPU memory consumption is directly proportional to the window size, thus a window size of 60, with an unrolling length of 200 steps, emerges as an optimal balance. As such, we typically maintain a window size to unrolling length ratio of around 1:3.

In our implementation, we employ a window size and unrolling length of (60, 200) for CIFAR-10 IPC1 and CUB-200, (80, 250) for CIFAR-10 IPC10, and (100, 300) for all other datasets.

\subsection{Other Architectures}

We further assessed our method across various architectures to demonstrate its universality. It is noteworthy that our approach is already effective across different widths of the convolutional networks (narrow and wide) we used. Additionally, we conducted tests using the standard ResNet-18, both  training it from scratch and  transferring from the distilled dataset. To our knowledge, we are the pioneers in applying direct distillation to a standard-sized network like ResNet-18. Direct training of ResNet-18 yields an accuracy of 52.7\% on CIFAR10 with IPC10, while the transfer from a 3-layer CNN results in 49.2\%. These transfer findings align with prior observations documented in \cite{deng2022remember}.

\subsection{Visualization}

\begin{figure}[htb]
    \centering
    \includegraphics[width=\linewidth]{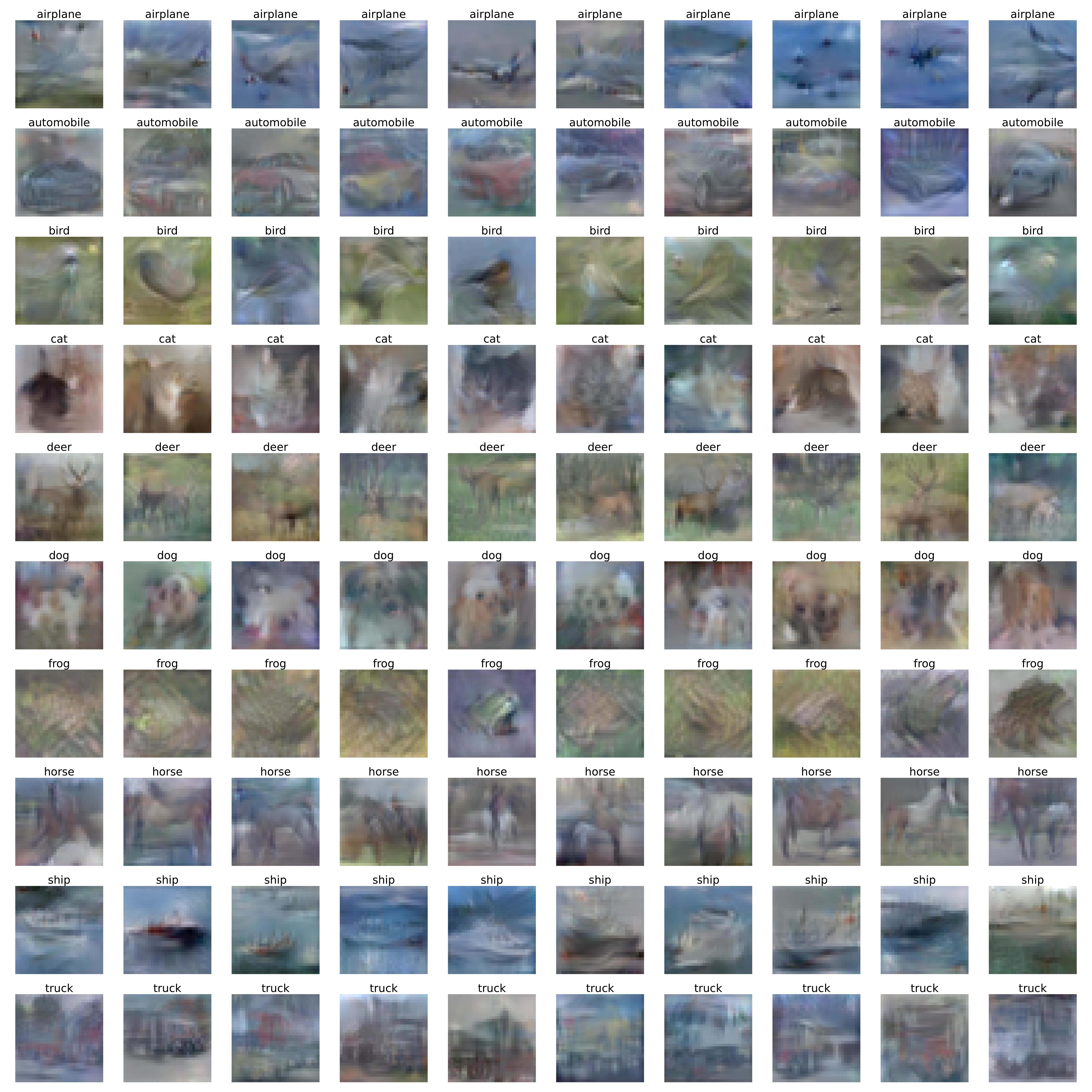}
    \caption{Visualization for RaT-BPTT standardly trained on CIFAR-10 with IPC10.}
    \label{fig:ipc_std}
\end{figure}

\begin{figure}[htb]
    \centering
    \includegraphics[width=\linewidth]{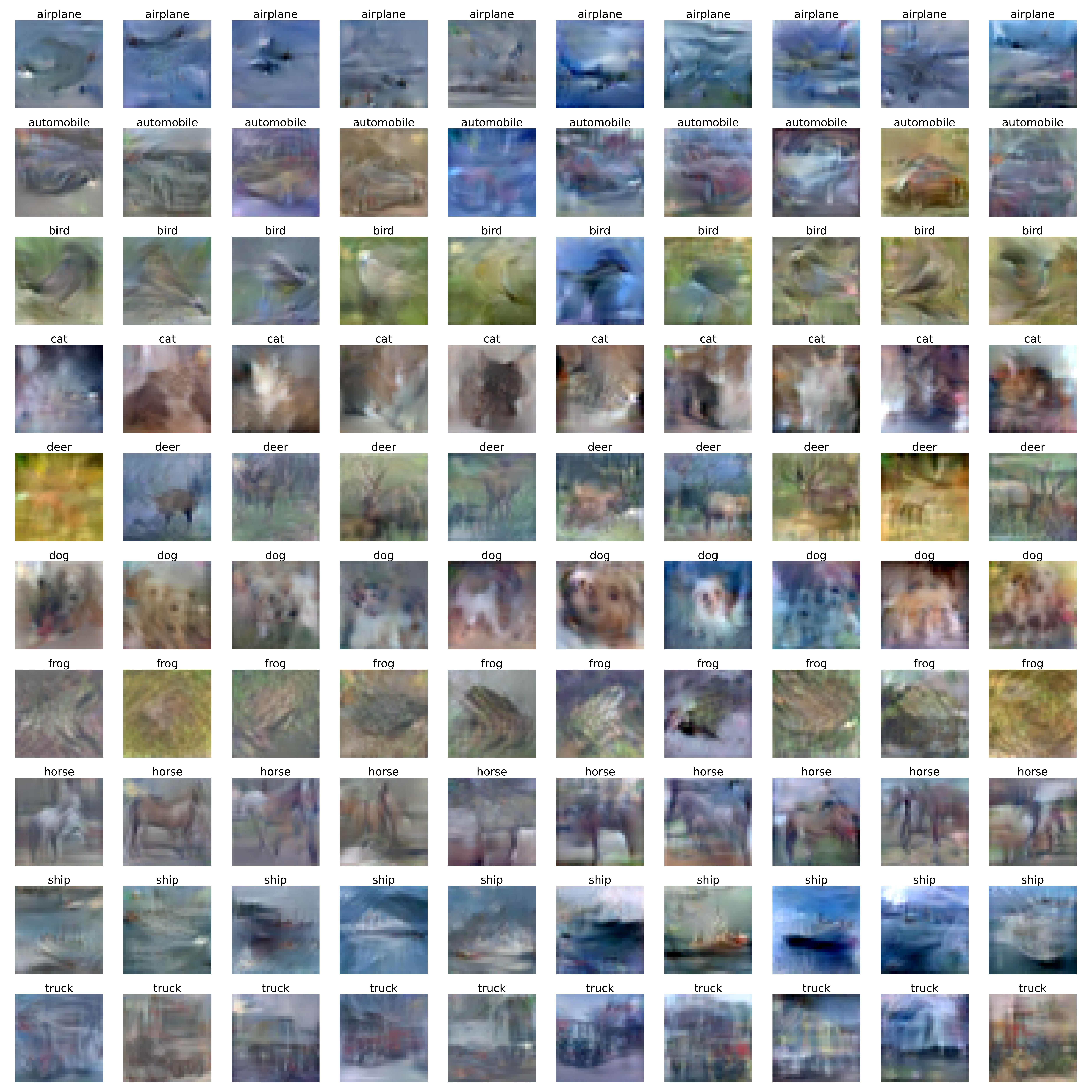}
    \caption{Visualization for RaT-BPTT with strong boosting (Boost-DD). CIFAR-10 with IPC10.}
    \label{fig:ipc_strong}
\end{figure}

\begin{figure}[htb]
    \centering
    \includegraphics[width=\linewidth]{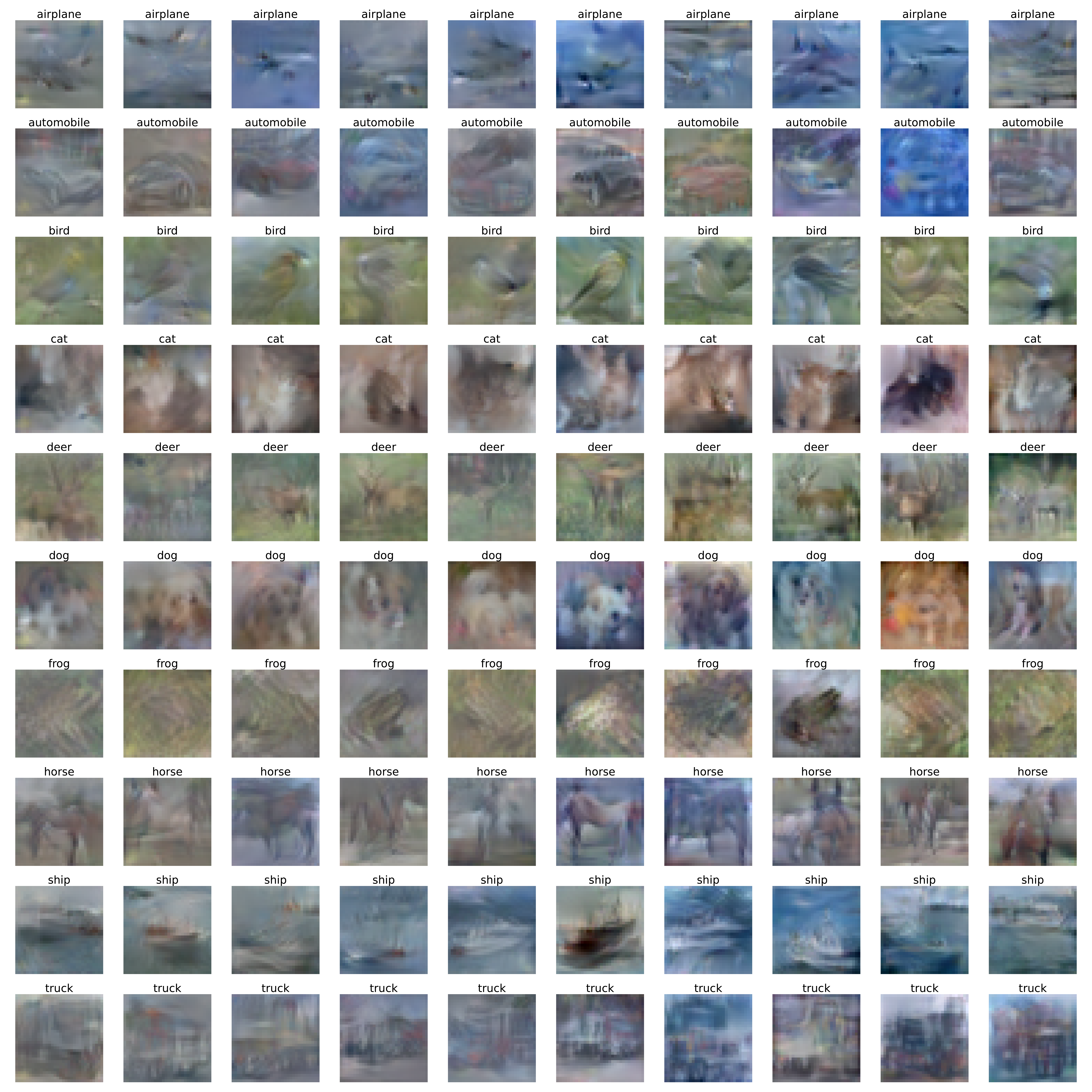}
    \caption{Visualization for RaT-BPTT with weak boosting (Boost-DD). CIFAR-10 with IPC10.}
    \label{fig:ipc_weak}
\end{figure}

We incorporate visualizations for IPC10 on CIFAR-10, representing standardly trained (Figure \ref{fig:ipc_std}), weakly boosted (Figure \ref{fig:ipc_weak}), and strongly boosted images (Figure \ref{fig:ipc_strong}). Upon inspection, the images from both boosted categories appear more diverse compared to their standard counterparts. 

\clearpage

\clearpage

\section{Hardness Analysis} \label{sec:hardness}

\subsection{Dissecting the Data}

% \begin{figure}[tb]
%     \begin{minipage}[t]{0.48\linewidth}
%         \centering
%         \includegraphics[width=1\linewidth]{figs/final/hardness.pdf}
%     \vspace{-12pt}
%     \caption{\textbf{Top:} Forgetting score~\cite{toneva2018forgetting} vs accuracy of dataset distillation (for various images per class (IPC)). ``Optimal'' indicates accuracy when training on the entire 
%     real data. \textbf{Bottom:} Histogram of the forgetting scores. Notice how score 0 (unforgettable) examples are distilled well (top), but harder
%     examples %(score > 0) 
%     are progressively harder to distill. \julia{Cut this figure}}
%     \label{fig:hardness-full}
%     \end{minipage}
%     \hfill
%     \begin{minipage}[t]{0.48\linewidth}
%         \centering
%         \includegraphics[width=\textwidth]{figs/final/quaprob.pdf}
%     \vspace{-10pt}
%     \caption{Test performance of hardness sampler versus standard for thresholds 1 and 4. We  start to sample at 25K steps since we believe it becomes more relevant in the later stages of distillation. Setting: CIFAR10, IPC50, RaT-BPTT. \julia{Move this figure down to D.2} }
%     \label{fig:sampler}
%     \end{minipage}
% \end{figure}

Now, we attempt to understand what is bottlenecking the gains when scaling up the distilled data from say IPC1 to IPC50 (see Table 1) by analyzing the performance of dataset distillation on examples that are easy or hard to learn, using a {\em hardness score} to stratify the data.
We first leverage the {\em forgetting score} \cite{toneva2018forgetting} as an empirical implementation of the hardness score as it characterizes the difficulty of learning a datapoint along the {\em training} trajectory.

\noindent\textbf{Forgetting score.} For each data point $x$ and a network trained for $T$ epochs, we say $x$ has a {\em forgetting event} at time $t$ if $x$ is correctly classified at time $t$ and misclassified at time $t+1$. %A {\em learning event} at epoch $t$ occurs if $x$ is misclassified at $t$ and correctly classified at $t+1$. 
The {\em forgetting score} of a data point is the sum of its forgetting events before time $T$, averaged over $10$ random initializations. \cite{toneva2018forgetting} show that based on the forgetting score a significant amount of training data can be omitted, without sacrificing test performance.

We first compute the forgetting score of the original network for each training data point. We then 
stratify the accuracy by forgetting score and compare it for the original network and the network trained on distilled data (Figure \ref{fig:hardness}). Notice that the highest boost in performance,
especially for the easy datapoints (score 0) comes from simply having
one image per class (IPC1). Further, one would have hoped
that increasing the number of images per class would help distill
more of the tail, enabling models to generalize
better to data points with larger forgetting scores. We notice
that this happens to some extent till a  score of 4,
but after that despite there being a lot of datapoints
with a larger hardness score (Figure~\ref{fig:hardness}, bottom) IPC50
seems to yield minimal marginal improvements. This suggests
that future works might benefit from focusing on how one
can distill datapoints with larger forgetting scores.

% \begin{comment}

\subsection{Hardness Sampler}

We present an initial approach to enhance learning on challenging examples through a "hardness sampler" that modifies the data batch distribution. Specifically, our objective is to enrich validation batches with more challenging examples, concentrating primarily on those lying mid-way between the extremes of very easy and very hard examples. This approach is inspired by the parabolic shape depicted in Figure \ref{fig:hardness}.

\begin{wrapfigure}{r}{0.4\textwidth}
%\begin{figure}[htb]
\vspace{-10pt}
    \centering
    \includegraphics[width=0.4\textwidth]{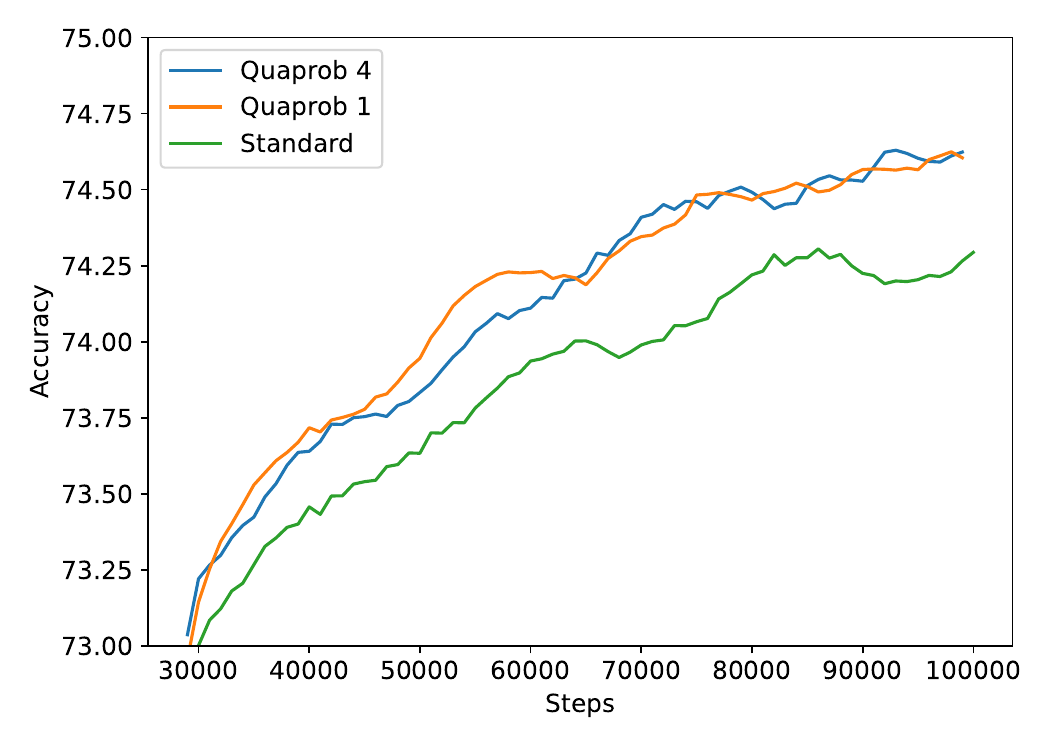}
    \vspace{-10pt}
    \caption{Test performance of hardness sampler versus standard for thresholds 1 and 4. We start to sample at 25K steps since we believe it becomes more relevant in the later stages of distillation. Setting: CIFAR10, IPC50, RaT-BPTT. }
    \label{fig:sampler}
%\end{figure}
\vspace{13pt}
\end{wrapfigure}

However, the calculation of the forgetting score is often computationally demanding and thus may not be practical for all applications, especially as part of the outer loop in dataset distillation. Moreover, Forgetting score has only been defined for and analysed on networks that are trained on the data that is being scored. To address these challenges, we propose an \textbf{adaptive hardness score} that is both efficient and versatile. This score is computed based on the disagreement in predictions \cite{feng2020transferred} across 8 randomly trained networks using the current distilled dataset. To stay relevant to the evolving challenges, this score is updated adaptively every 50 epochs.

Based on this adaptive hardness score, we down-weight the easiest and hardest examples by giving the following weight $w$ to a sample $x$ with hardness score 
$HS(x) \in \{0,\ldots,8\}$:
\begin{equation*}
    w(x)=thr+abs(HS(x)-4),
\end{equation*}
where $thr$ is a threshold which we set to either $1$ or $4$. For each update of the meta-gradient we sample from a training data point $x$ proportional to $w(x)$, which upweights medium-hard examples the most. Figure~\ref{fig:sampler} demonstrates a notable performance improvement for the IPC50 distillation on CIFAR10, and hints that this direction might be fruitful for future work to pursue.

%\end{table*}